\documentclass[parskip=full,12pt]{article}

\usepackage{amssymb}
\usepackage{amsthm}
\usepackage{amsmath}

\usepackage{float,booktabs}
\usepackage{float} 
\usepackage{graphicx}
\usepackage{natbib}
\usepackage{url}
\usepackage{xcolor}
\usepackage{soul,color,colortbl}
\usepackage{array,multirow}
\usepackage{cancel}
\usepackage[top=1in, bottom=1in, left=0.9in, right=0.9in]{geometry}
\usepackage[linesnumbered,ruled]{algorithm2e}
\usepackage{tikz}
\usetikzlibrary{matrix,chains,positioning,decorations.pathreplacing,arrows}
\usepackage[labelformat=simple]{subcaption}
\usepackage{bm}
\usepackage[linesnumbered,ruled]{algorithm2e}
\usepackage{caption}

\DeclareRobustCommand{\1}[1]{\text{\usefont{U}{bbold}{m}{n}1}_{#1}}
\usepackage{soul}

\def\spacingset#1{\renewcommand{\baselinestretch}
{#1}\small\normalsize} \spacingset{1}

\newcolumntype{P}[1]{>{\centering\arraybackslash}p{#1}}
\newcommand*{\myfont}{\fontfamily{lmss}\selectfont}
\DeclareTextFontCommand{\textpython}{\myfont}

\DeclareCaptionLabelFormat{boldnumber}{\bothIfFirst{#1}{~}\textbf{#2}}
\title{\textbf{Synthetic Dataset Generation of Driver Telematics}}

\author{Banghee So\thanks{Department of Mathematics, University of Connecticut, 341 Mansfield Road, Storrs, CT, 06269-1009, USA. Email: \texttt{banghee.so@uconn.edu}.} 
\and Jean-Philippe Boucher\thanks{D\'{e}partement de Math\'{e}matiques, Universit\'{e} du Qu\'{e}bec \`{a} Montr\'{e}al, 201, Avenue du Pr\'{e}sident-Kennedy, Montr\'{e}al, Qu\'{e}bec, H2X 3Y7, Canada. Email: \texttt{boucher.jean-philippe@uqam.ca}.}
\and Emiliano A. Valdez\thanks{Corresponding author; Department of Mathematics, University of Connecticut, 341 Mansfield Road, Storrs, CT, 06269-1009, USA. Email: \texttt{emiliano.valdez@uconn.edu}.}}

\begin{document}

\maketitle

\begin{abstract}
This article describes techniques employed in the production of a synthetic dataset of driver telematics emulated from a similar real insurance dataset. The synthetic dataset generated has 100,000 policies that included observations about driver's claims experience together with associated classical risk variables and telematics-related variables. This work is aimed to produce a resource that can be used to advance models to assess risks for usage-based insurance. It follows a three-stage process using machine learning algorithms.  The first stage is simulating values for the number of claims as multiple binary classifications applying feedforward neural networks. The second stage is simulating values for aggregated amount of claims as regression using feedforward neural networks, with number of claims included in the set of feature variables. In the final stage, a synthetic portfolio of the space of feature variables is generated applying an extended \texttt{SMOTE} algorithm. The resulting dataset is evaluated by comparing the synthetic and real datasets when Poisson and gamma regression models are fitted to the respective data. Other visualization and data summarization produce remarkable similar statistics between the two datasets.  We hope that researchers interested in obtaining telematics datasets to calibrate models or learning algorithms will find our work valuable.

\vspace{1cm}

\noindent \textbf{Keywords}: Bayesian optimization, Gaussian process, Neural network, \texttt{SMOTE}, Usage-based insurance (UBI), Vehicle telematics.

\end{abstract}

\newpage

\section{Background} \label{sec:intro}

Usage-based insurance (UBI) is a recent innovative product in the insurance industry that exploits the use and access of improved technology. It is a type of automobile insurance policy where the cost of insurance is directly linked to the usage of the automobile. With the help of telematics device or mobile app, auto insurers are able to track and monitor mileage, speed, acceleration, and other driving-related data. This data transmission allows insurers to later store information for monitoring driving behavior and subsequently, for risk assessment purposes.

According to the Oxford dictionary, telematics refers to ``the use or study of technology that allows information to be sent over long distances using computers.'' Its origin can be traced back to the French word, t\'{e}l\'{e}matique,  combining the words ``telecommunications'' and ``computing science.'' There is a growing list of applications of telematics in various industries, and it is most prominently used in the insurance industry. The infrastructure offered by health telematics allows for access to healthcare that helps reduce costs while optimizing quality of patient care. The installation of a smart home system with alarms that remotely monitor home security can drastically reduce cost of homeowners insurance. In auto insurance, a plug-in device, an integrated equipment installed by car manufacturers, or a mobile application can be used to directly monitor cars thereby allowing insurers to more closely align driving behaviors with insurance premium rates through UBI. It was said in \citet{cipr2015} that Progressive Insurance Company, in collaboration with General Motors, offered the first such UBI in the early 2000s that offered premium discounts linked to monitoring of driving activities and behavior. With agreement of the driver, a tracking device was installed in the vehicle to collect information through GPS technology. Subsequently, with even further advances in technology, different forms of UBI have emerged that include, for example, Pay-as-you-Drive (PAYD), Pay-how-you-Drive (PHYD), Pay-as-you-Drive-as-you-Save (PAYDAYS), Pay-per-mile, and Pay-as-you-Go (PASG).

The variations in UBI programs generally fall into two broad categories: how you drive and how far you drive. In the first category, insurers track data, such as the changes in your speed, how fast you are driving as you make a left or right turn, the day of the week you drive, and the time of day you drive, that reflects your driving maneuvering behavior. In the second category, insurers track data that is related to your driving mileage, essentially the distance you travel in miles or kilometers. It is interesting to note that, even prior to development of telematics, \citet{butler1993carmile} have suggested the use of cents-per-mile premium rating for auto insurance. See also \citet{denuit2007} for an early discussion of the development of PAYD auto pricing.

\subsection{Literature} \label{sec:lit1}

The actuarial implications of usage-based insurance for fair risk classification and a more equitable premium rating are relevant; this is reflected in the growth in the literature on telematics in actuarial science and insurance. Many of the research on telematics have found the additional value of information derived from telematics to provide improved claims predictions, risk classification, and premium assessments. \citet{husnjak2015UBI} provides a very nice overview of the architecture and pricing paradigms employed by various telematics programs around the world.

Table \ref{tab:RO} provides an overview of the literature in actuarial science and insurance, with an outline of the work describing the data source, the period of observation with sample size, the analytical techniques employed, and a brief summary of the research findings. For example, the early work of \citet{ayuso2014time} examines a comparison of the driving behaviors between novice and experienced young drivers, those aged below 30, with PAYD policies. The analysis is based on a sample of 15,940 young drivers with PAYD policies in 2009 drawn from a leading Spanish insurance company. The work of \citet{guillen2020telem} demonstrates how the additional information drawn from telematics can help predict near-miss events. The analysis is based on a pilot study of drivers from Greece in 2017 who agreed to participate in a telematics program.

\begin{table}[H]
\centering
\resizebox{!}{6.5cm}{
\begin{tabular}{lp{4.45cm}lp{1.5cm}lp{9.5cm}} \\
\toprule
Data source	& Reference	&Sample&Period	&Analytical techniques &Research synthesis \\
\midrule
\vtop{\hbox{\strut Belgium}}&\citet{verbelen2018telem} & \vtop{\hbox{\strut10,406 drivers}\hbox{\strut (33,259 obs.)}}&2010-2014&\vtop{\hbox{\strut Poisson GAM,}\hbox{\strut Negative binomial GAM}}&Shows that the presence of telematics variables are better important predictors of driving habits \\
\midrule
\vtop{\hbox{\strut Canada}}&\citet{so2020cost}&71,875 obs.&2013-2016&\vtop{\hbox{\strut Adaboost,}\hbox{\strut SAMME.C2}}&Demonstrates telematics information improves the accuracy of claims frequency prediction with a new boosting algorithm \\
\midrule
China&\citet{gao2019telfreq}&1,478 drivers&2014.01-2017.06&Poisson GAM&Shows the relevance of telematics covariates extracted from speed-acceleration heatmaps in a claim frequency model \\
\midrule
\vtop{\hbox{\strut Europe}}&\citet{baecke2017risk}&\vtop{\hbox{\strut 6,984 drivers}\hbox{\strut ($<$ age 30)}}&2011-2015&  \vtop{\hbox{\strut Logistic regression,}\hbox{\strut Random forests,}\hbox{\strut Neural networks}}&Illustrates the importance of telematics variables for pricing UBI products and shows that as few as three months of data may already be enough to obtain efficient risk estimates \\
\midrule
Greece&\citet{guillen2020telem}& \vtop{\hbox{\strut 157 drivers}\hbox{\strut (1,225 obs.)}}&2016- 2017&Negative binomial reg.&Demonstrates how the information drawn from telematics can help predict near-miss events \\
\midrule
\vtop{\hbox{\strut Japan}}&\citet{osafune2017acc}&809 drivers&2013.12-2015.02& Support Vector Machines & Investigates accident risk indices that statistically separate safe and risky drivers \\
\midrule
\vtop{\hbox{\strut Spain}}&\citet{ayuso2014time}& \vtop{\hbox{\strut 15,940 drivers}\hbox{\strut ($<$ age 30)}} & 2009-2011&Weibull regression &Compares driving behaviors of novice and experienced young drivers with PAYD policies \\
& \citet{ayuso2016gender}& \vtop{\hbox{\strut 8,198 drivers}\hbox{\strut ($<$ age 30)}}&2009-2011&Weibull regression&Determines the use of gender becomes irrelevant in the presence of sufficient telematics information \\
& \citet{boucher2017gam}&71,489 obs. &2011&Poisson GAM&Offers the benefits of using generalized additive models (GAM) to gain additional insights as to how premiums can be more dynamically assessed with telematics information \\
&\citet{guillen2019telem}&\vtop{\hbox{\strut 25,014 drivers}\hbox{\strut ($<$ age 40)}} &2011&Zero-inflated Poisson &Investigates how telematics information helps explain part of the occurrence of zero accidents not typically accounted by traditional risk factors \\
&\citet{ayuso2019auto}&\vtop{\hbox{\strut 25,014 drivers}\hbox{\strut ($<$ age 40)}} &2011&Poisson regression&Incorporates information drawn from telematics metrics into classical frequency model for tariff determination \\
&\citet{perez2019quantile} &\vtop{\hbox{\strut 9,614 drivers}\hbox{\strut ($<$ age 35)}} &2010&Quantile regression&Demonstrates that the use of quantile regression allows for better identification of factors associated with risky drivers \\
&\citet{pesantez2019xgboost}&\vtop{\hbox{\strut 2,767 drivers}\hbox{\strut ($<$ age 30)}}&2011&XGBoost&Examines and compares the performance of XGBoost algorithm against the traditional logistic regression\\
\bottomrule
\end{tabular}}
\caption{An overview of the literature.} \label{tab:RO}
\end{table}

\subsection{Motivation} \label{sec:lit2}

Here in this article, we provide the details of the procedures employed in the production of a synthetic dataset of driver telematics. This synthetic dataset was generated to imitate the intricate characteristics of a similar real insurance dataset; the intent is not to reproduce nor replicate the original characteristics in order to preserve the privacy that may be alluded from the original source. In the final synthetic dataset generated, we produced 100,000 policies that included observations about driver's information and claims experience (number of claims and aggregated amount of claims) together with associated classical risk variables and telematics-related variables. As previously discussed, an increasingly popular auto insurance product innovation is usage-based insurance (UBI) where a tracking device or a mobile app is installed to monitor insured driving behaviors. Such monitoring is an attempt of the industry to link risk premiums assessed with observable variables that are more directly tied to driving behaviors. While such monitoring may be engineered more frequently than that reproduced or implied  in our synthetic dataset, the dataset is in aggregated or summarized form assumed to be observed over a certain period of time and can be used for research purposes of performing risk analysis of UBI products. For the academic researcher, the dataset can be used to calibrate advances in actuarial and risk assessment modeling. On the other hand, the practitioner may find the data useful for market research purposes where for instance, an insurer is intending to penetrate the UBI market.

In the actuarial and insurance community as driven by industry need that is facilitated with computing technology advancement, there is a continuing growth of the need for data analytics to perform risk assessment with high accuracy and efficiency. Such exercise involves the construction, calibration, and testing of statistical learning models, which in turn, requires the accessibility of big and diverse data with meaningful information. Access to such data can be prohibitively difficult, understandably so because several insurers are reluctant to provide data to researchers for concerns of privacy.

This drives a continuing interest and demand for synthetic data that can be used to perform data and predictive analytics. This growth is being addressed in the academic community. To illustrate, the work of \citet{gan2017va} and \citet{gan2018nested} created synthetic datasets of large portfolios of variable annuity products so that different metamodeling techniques can be constructed and tested. Such techniques have the potential benefits of addressing the intensive computational issues associated with Monte Carlo techniques typically common in practice. Metamodels have the added benefits of drastically reducing computational times and thereby providing a more rapid response to risk management when market forces drive the values of these portfolios. \citet{gabrielli2018indclms} developed a stochastic simulation machinery to reproduce a synthetic dataset that is ``realistic'' and reflects real insurance claims dataset; the intention is for analysts and researchers to have access to a large data in order to develop and test individual claims reserving models. Our paper intends to continue this trend of supporting researchers by providing them with a synthetic dataset to allow them to calibrate advancing models. More specifically, we build the data generating process to produce an imitation of the real telematics data. The procedure initially constructs two neural networks, which emulates the number of claims and aggregated amount of claims that can be drawn from real data. We then generate 100,000 synthetic observations with features using extended version of \texttt{SMOTE}. Inserting the synthetic observations into two neural networks, we are able to produce the complete portfolio with the synthetic number of claims and aggregated amount of claims.    

The rest of this paper has been structured as follows. Section \ref{sec:work} describes the machine learning algorithms used to perform the data generation. Section \ref{sec:data} provides a description of all the variables included in the synthetic datafile. Section \ref{sec:datagen} provides the details of the data generation process using the feedforward neural networks and the extended \texttt{SMOTE}. This section also provides the comparison of the real data and the synthetically generated data when Poisson and gamma regression models are used. We conclude in Section \ref{sec:conclude}.

\section{Related work} \label{sec:work}

This section briefly explains two popular machine learning algorithms that we employed to generate the telematics synthetic dataset. The first algorithm is the extended \texttt{SMOTE}, Synthetic Minority Oversampling Technique. This procedure is used to generate the classical and telematics predictor variables in the dataset. The second algorithm is the feedforward neural network. This is used to generate the corresponding response variables that describe number of claims and the aggregated amount of claims.

\subsection{Extended SMOTE} \label{sub:smote}

Developed by \citet{chawla2002smote}, the Synthetic Minority Oversampling Technique (\texttt{SMOTE}) is originally intended to address classification datasets with severe class imbalances. The procedure is to augment the data to oversample observations for the minority class and this is accomplished by selecting samples that are within the neighborhood in the feature space. First, we choose a minority class and then we obtain its $K$-nearest neighbors, where $K$ is typically set to 5. All $K$ neighbors should be minority instances. Then, one of these $K$ neighbor instances  are  randomly  chosen  to  compute new  instances by interpolation. The interpolation is performed by computing the difference between the minority class instance under consideration and the selected neighbor taken.  This difference is multiplied by a random number uniformly drawn between 0 and 1, and the resulting instance is added to the considered minority class. In effect, this procedure does not duplicate observations, however, the interpolation causes the selection of a random point along the ``line segment'' between the features (\citet{fernandez2018smote}).

This principle of \texttt{SMOTE} for creating synthetic data points from minority class is employed and adopted in this paper with a minor adjustment. In our data generation, we applied it to generate predictor variables based on the entire feature space of the original or real dataset. The one minor adjustment we used is to tweak the interpolation by randomly drawing a number from a $U$-shaped distribution, rather than a uniform distribution, between 0 and 1. This mechanism has the resulting effect of maintaining the characteristic of the original or real dataset with small possibility of duplication. In particular, we are able to capture characteristics of observations that may be considered unusual or outliers. Further description of synthetically generated portfolio is given in Section \ref{sub:pro3}.

\subsection{Feedforward neural network} \label{sub:NN}

Loosely modeled after the idea of neurons that form the human brain, neural network consists of a set of algorithms for doing machine learning in order to cleverly recognize patterns. Neural networks are indeed very versatile as they can be used for addressing inquiries that are considered either supervised or unsupervised learning; this set of algorithms has grown in popularity as the method continues to provide strong evidence of its ability to produce predictions with high accuracy. A number of research using neural networks has been published in the actuarial and insurance literature. \citet{wuthrich2019bias} showed that the biased estimation issue resulting from use of neural networks with early stopping rule can be diminished using shrinkage version of regularization. \citet{yan2020improved} used backpropagation (BP) neural network optimized by an improved adaptive genetic algorithm to build car insurance fraud detection model. Additional research has revealed the benefits and advantages of neural networks applied to various models for insurance pricing, fraud detection, and underwriting. Among these include, but are not limited to, \citet{viaene2005auto}, \citet{dalkilic2009neural}, \citet{ibiwoye2012ann}, and \citet{kiermayer2020grouping}.

The idea of neural networks can be attributed to the early work of \citet{mcculloch1943NN}. A neural network (NN) consists of several processing nodes, referred to as neurons, considered to be simple yet densely interconnected. Each neuron produces a sequence of real-valued activations triggered by a so-called activation function, and these neurons are organized into layers to form a network. The activation function plays a crucial role in the output of the model, affecting its predictive accuracy, computational efficiency of learning a model, and convergence. There are several types of neural network activation functions, and we choose just a few of them for our purpose.

Neural network algorithms have the tendency to be complex and to overfit the training dataset. Because of this model complexity, they are often referred to as black-box as it becomes difficult sometimes to draw practical insights into the learning mechanisms employed. Part of this problem has to do with the large number of parameters and the resulting non-linearity of the activation functions. However, these disadvantageous features of the model may be beneficial for the purpose of our data generation. For instance, the overfitting may help us build a model with high accuracy and precision so that we produce a synthetic portfolio that mimics the characteristics of the portfolio derived from the real dataset.

For feedforward neural networks, signals are more straightforward because they are allowed to go in one direction only: from input to output (\citet{Goodfellow2016}). In effect, the output from any layer does not directly affect that same layer so that the effect is that there are no resulting feedback loops. In contrast, for recurrent neural networks, signals can travel in both directions so that feedback loops may be introduced in the network. Although considered more powerful, computations within recurrent neural networks are much more complicated than those within feedforward neural networks. As later described in the paper, we fit two simulations using the feedforward neural network. 

\begin{figure}[h!]
\centering
\includegraphics[scale=0.4325]{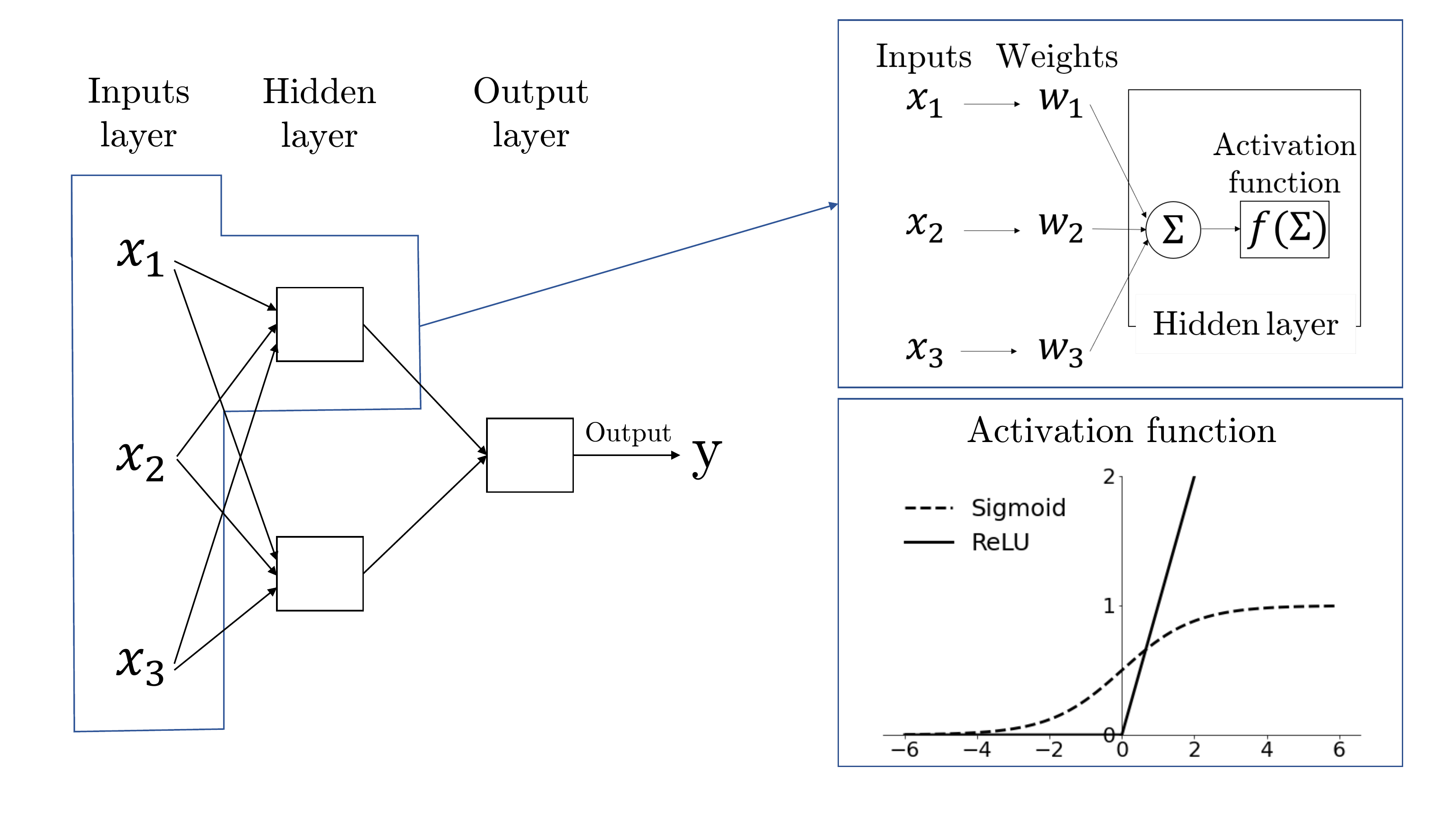} 
\caption{Architecture of a feedforward neural network.} \label{fig1:NNplot} 
\end{figure}

Figure \ref{fig1:NNplot} displays a sample architecture of a feedforward neural network, together with the type of activation functions considered in this article. In this case,  it becomes apparent how the information flows only from the input to the output. The graphs described in Figure \ref{fig1:NNplot} has three feature variables as the input, one hidden layer, two nodes for the hidden layer, and the response variable $y$ as the resulting output. The activation function ($f$) is responsible for converting weighted sum of previous node values ($\sum$) into a node value of that layer. Representative activation functions are sigmoid and Rectified Linear Unit (\texttt{ReLU}) functions as seen in the bottom left of Figure \ref{fig1:NNplot}. The sigmoid is used as an activation function in neural network that converts any real-valued sample to a probability range between 0 and 1. It is this property that the neural network can be used as binary classifier. On the other hand, the \texttt{ReLU} function is a piecewise linear function that gives the input directly as output, if positive, and zero as output, otherwise. This function is often the default function for many neural network algorithms because it is believed to train the model with ease and with outstanding performance.

In the feedforward neural network, parameters are the weights ($w_i$) of connections between layers. Hyperparameters are the values to determine the architecture of the neural network model, which include, among others, the number of layers, the number of nodes in each layer, activation functions, and parameters used for optimizer (e.g., Stochastic Gradient Descent (\texttt{SGD}) learning rate). Parameters can be learned from the data using a loss optimizer. However, hyperparameters still must be predetermined prior to the learning process and, in many cases, these decisions depend on the judgment of the analyst or the user. The work of \citet{hornik1989universal} proved that standard multi-layer feedforward networks are capable of approximating any measurable function, and thus is called the universal approximator. This implies that any lack of success in applications must arise from inadequate learning, insufficient numbers of hidden units, or the lack of a deterministic relationship between input and target. Hyperparameters may be more essential in deep learning to be able to yield satisfactory output.

We found that a number of research done in neural networks focused on introducing the algorithms for optimizing hyperparameters values. Some of the frequently used searching strategies are grid search, random search (\citet{bergstra2012randomsearch}), and sequential model-based optimization (\citet{bergstra2011smbo}). This line of work on hyperparameters is presently a very active field of research that includes, for example, hyperparameters in parameter learning process (e.g., \citet{thiede2019gradhyper}, \citet{franceschi2017gradhyper}, and \citet{maclaurin2015gradhyper}). However, the methods proposed in the current literature are relatively new and not mature enough to be used in practical real world problems. The simple and widely used optimization algorithms are the grid search and the random search. The grid search, on one hand, is the method to discretize the search space of each hyperparameter and based on the Cartesian products, to discretize the total search space of hyperparameters. Then, after learning for each set of the hyperparameters, we select the best at the end. It is intuitive and easy to apply but it does not take into account relative feature importance, and therefore is considered ineffective and extremely time-consuming. This method is also severely influenced by the curse of dimensionality as the number of hyperparameters increase. In the random search, on the other hand, hyperparameters are randomly sampled. \citet{bergstra2012randomsearch} showed that the random search, as compared to the grid search, is particularly effective, especially when dealing with relative feature importance. However, since the next trial set of hyperparameters are not chosen based on previous results, it is also time-consuming especially when it involves a large number of hyperparameters, thereby suffering from the same curse of dimensionality as the grid search.

To optimize hyperparameters, we find that one of the most powerful strategies is the sequential model-based optimization, also sometimes referred to as Bayesian optimization. The following set of hyperparameters are determined based on the result of previous sets of hyperparameters.  \citet{bergstra2011smbo} and \citet{snoek2012practical} showed that sequential model-based optimization outperforms both grid and random searches. Sequential model-based optimization constructs a probabilistic surrogate model to define the posterior distribution over unknown black box function (loss function). The posterior distribution is developed based on conditioning on the previous evaluations and a proxy optimization is performed to seek the next location to evaluate. For the proxy optimization, the acquisition function is computed based on the posterior distribution and has the highest value at the location having the highest probability of the lowest loss function; this point becomes the next location. Most commonly, Gaussian process is used as surrogate model because of their flexibility, well-calibrated uncertainty, and analytic properties (\citet{murugan2017hyperparameters}). Thus, we use the Gaussian process as the hyperparameter tuning algorithm.  

Another important decision, which may affect the time efficiency and performance of the neural network model, is to choose the optimizer. The optimizer refers to an algorithm used to update parameters of model in order to reduce the losses. Neural network is not a convex optimization. For this reason, in the training process, it could fall into the minimum of local part and the convergence rate could be too small leading to the learning process unfinished for days (\citet{li2012bp}). To address this issue, diverse optimizers have been suggested: Gradient Descent, Stochastic Gradient Descent, Mini-Batch Gradient Descent, Momentum, \texttt{AdaGrad} (\citet{duchi2011adaptive}), \texttt{RAMSProp} (\citet{hinton2012neural}), \texttt{Adam} (\citet{kingma2014adam}) and others (\cite{ruder2016overview}). The \texttt{Adam} optimization is an efficient stochastic optimization that has been suggested and it combines the advantages of two popular methods: \texttt{AdaGrad}, which works well with sparse gradients, and RMSProp, which has an excellent performance in on-line and non-stationary settings. Recent works by \citet{zhang2019adam}, \citet{peng2018chemical}, \citet{bansal2016ask} and \citet{arik2017convolutional} have presented and proven that \texttt{Adam} optimizer provides better performance than others in terms of both theoretical and practical perspectives. Therefore in this paper, we use \texttt{Adam} as the optimizer in our neural network simulations. 

\section{The synthetic output: file description} \label{sec:data}

For our portfolio emulation, we based it on a real dataset acquired from a Canadian-based insurer, which offered a UBI program that was launched in 2013, to its automobile insurance policyholders. The observation period was for years between 2013 and 2016, with over 70,000 policies observed for which the dataset drawn is pre-engineered for training a statistical model for predictive purposes. See also \citet{so2020cost}.

\begin{table}[h!]
\centering
\resizebox{!}{6cm}{
\begin{tabular}{lll} \\
\toprule
Type & Variable  & Description \\
\midrule
Traditional & \texttt{Duration} & Duration of the insurance coverage of a given policy, in days\\ 
 & \texttt{Insured.age}  & Age of insured driver, in years \\
 & \texttt{Insured.sex}  & Sex of insured driver (Male/Female) \\
 & \texttt{Car.age}  & Age of vehicle, in years \\
 & \texttt{Marital}  & Marital status (Single/Married) \\
 & \texttt{Car.use} & Use of vehicle: Private, Commute, Farmer, Commercial  \\
 & \texttt{Credit.score}  & Credit score of insured driver \\
 & \texttt{Region}  & Type of region where driver lives: rural, urban \\
 & \texttt{Annual.miles.drive}  & Annual miles expected to be driven declared by driver \\
 & \texttt{Years.noclaims}  & Number of years without any claims\\
 & \texttt{Territory}  & Territorial location of vehicle \\
\midrule
Telematics & \texttt{Annual.pct.driven}  & Annualized percentage of time on the road \\
 &\texttt{Total.miles.driven}  &Total distance driven in miles \\
 &\texttt{Pct.drive.xxx}	&Percent of driving day xxx of the week: mon/tue/…/sun\\
 &\texttt{Pct.drive.xhrs}&Percent vehicle driven within x hrs: 2hrs/3hrs/4hrs\\
 &\texttt{Pct.drive.xxx}	&Percent vehicle driven during xxx: wkday/wkend\\
 &\texttt{Pct.drive.rushxx	}&Percent of driving during xx rush hours: am/pm\\
 &\texttt{Avgdays.week}	&Mean number of days used per week\\
 &\texttt{Accel.xxmiles}	&Number of sudden acceleration 6/8/9/\ldots/14 mph/s per 1000miles\\
 &\texttt{Brake.xxmiles}	&Number of sudden brakes 6/8/9/\ldots/14 mph/s per 1000miles\\
 &\texttt{Left.turn.intensityxx}	&Number of left turn per 1000miles with intensity 08/09/10/11/12\\
 &\texttt{Right.turn.intensityxx} &Number of right turn per 1000miles with intensity 08/09/10/11/12\\
\midrule
Response & \texttt{NB\_Claim}  & Number of claims during observation \\
 & \texttt{AMT\_Claim}  & Aggregated amount of claims during observation \\
\bottomrule
\end{tabular}}
\caption{Variable names and descriptions.} \label{tab:VD}
\end{table}

We generated a synthetic portfolio of 100,000 policies. Table \ref{tab:VD} provides the types, names, definitions or brief description of the various variables in the resulting datafile, which can be found in 
\begin{center}
\url{http://www2.math.uconn.edu/~valdez/data.html}.
\end{center}
The synthetic datafile contains a total of 52 variables, which can be categorized into three main groups: (a) 11 traditional features such as policy duration, age and sex of driver, (b) 39 telematics features including total miles driven, number of sudden breaks or sudden accelerations, and (3) 2 response variables describing number of claims and aggregated amount of claims.

Additional specific information of the variables in the datafile is presented below:
\begin{itemize}
\itemsep0em
\item \texttt{Duration} is the period that policyholder is insured in days, with values in [22,366].
\item \texttt{Insured.age} is the age of insured driver in integral years, with values in [16,103].
\item \texttt{Car.age} is the age of vehicle, with values in [-2,20]. Negative values are rare but are possible as buying a newer model can be up to two years in advance. 
\item \texttt{Years.noclaims} is the number of years without any claims, with values in [0, 79] and always less than \texttt{Insured.age}.
\item \texttt{Territory} is the territorial location code of vehicle, which has 55 labels in \{11,12,13,$\cdots$,91\}.
\item \texttt{Annual.pct.driven} is the number of day a policyholder uses vehicle divided by 365, with values in [0,1.1].
\item \texttt{Pct.drive.mon}, $\cdots$, \texttt{Pct.drive.sun} are compositional variables meaning that the sum of seven (days of the week) variables is 100\%. 
\item \texttt{Pct.drive.wkday} and \texttt{Pct.drive.wkend} are clearly compositional variables too. 
\item \texttt{NB\_Claim} refers to the number of claims, with values in \{0,1,2,3\}; 95.72\% observations with zero claim, 4.06\% with exactly one claim, and merely 0.20\% with two claim and 0.01\% with three claim. Real NB\_Claim has the following proportions; zero claim: 95.60\%, one claim: 4.19\%, two claim: 0.20\%, three claim: 0.007\%.
\item \texttt{AMT\_Claim} is the aggregated amount of claims, with values in [0, 138766.5]. Summary statistics of synthetic and real data is shown in Table \ref{tab:summ1}.
\end{itemize}

Table \ref{tab:summ1} provides an interesting comparison of the summary statistics of the aggregated amount of claims derived from the synthetic datafile and compared to the real dataset, broken down by the number of claims from the synthetic dataset. First, we observe that we do not exactly replicate the statistics, a good indication that we have done a good job of reconstructing a portfolio based on the real dataset with very little indication of reproducing nor replicating the exact data. Second, these statistics show that we are able to preserve much of the characteristics of the original dataset according to the spread and depth of observations we have as described in this table. To illustrate, among those with exactly 2 claims, the average amount of claim in the synthetic file is 8960 and it is 8643 in the real dataset; the median is 7034 in the synthetic file while it is 5148 in the real data. The respective standard deviations, which give a sense of how dispersed the values are from the mean, are 9554 and 10924. We shall be able to compare more of these intricacies when we evaluate the quality of the reproduction by giving more details of this type of comparisons.

\begin{table}[h!]
\centering
\resizebox{!}{1.4cm}{
\begin{tabular}{l|c||rrrrrrr} \cline{2-9}
Synthetic & NB\_Claim & Mean & Std Dev & Min & Q1 & Median & Q3 & Max \\ \hline
AMT\_Claim & 0 & 0 & 0 & 0 & 0 & 0 & 0 & 0 \\ \cline{2-9}
 & 1 & 4062 & 6767 & 0 & 670 & 2191 & 4776 &	138767 \\ \cline{2-9}
 & 2 & 8960 & 9554 &	0 & 2350 & 7034 & 11225 & 56780 \\ \cline{2-9}
 & 3 & 5437 & 2314 & 2896 & 3620 & 5372 & 5698 & 9743 \\ \hline
\end{tabular}}

\vspace{1 cm}

\resizebox{!}{1.4cm}{
\begin{tabular}{l|c||rrrrrrr} \cline{2-9}
Real & NB\_Claim & Mean & Std Dev & Min & Q1 & Median & Q3 & Max \\ \hline
AMT\_Claim & 0 & 0 & 0 & 0 & 0 & 0 & 0 & 0 \\  \cline{2-9}
 & 1 & 4646 & 8387 & 0& 659& 2238& 5140& 145153 \\ \cline{2-9}
 & 2 & 8643 & 10920 & 0 & 1739 & 5184 & 11082 & 62259 \\ \cline{2-9}
 & 3 & 5682 & 2079 & 3253 & 4540 & 5416 & 5773 & 9521 \\ \hline
\end{tabular}} 
\caption{Summary statistics of AMT\_Claim based on synthetic NB\_Claim: Synthetic vs Real.} \label{tab:summ1}
\end{table}

As we said earlier, we reproduced 52 variables and the data types are summarized in Table \ref{tab:TV}. The NB\_Claim variables can be treated as integer-valued or a classification or categorical variable, with 0 category as those considered to be least risky drivers who thus far have zero claim frequency history. The percentage variables are those with values between 0 and 100\%. Compositional variables are less frequently described in insurance datasets but are increasingly becoming more important for telematics related variables. Compositional variables refer to a class or groups of variables that are commonly presented as percentages or proportions that describe parts of some whole. The total sum of these parts are typically constraint to be some fixed constant such as 100\% of the whole. A clear example in our dataset are the variables  \texttt{Pct.drive.wkday} and \texttt{Pct.drive.wkend}, for which respectively, are the percentages of times spent driving during the weekdays and during the weekends. For instance, if each of these are 50\%, then half of the time that the individual is driving on the road is done during the day of the week (Monday through Friday) while the other half is done during the weekend (Saturday and Sunday). See \citet{so2020cost} and \citet{verbelen2018telem}.

\begin{table}[H]
\centering
\resizebox{!}{3.75cm}{
\begin{tabular}{llll} \\
\hline \hline 
Category  & Continuous/Integer & Percentage &  Compositional \\
\midrule
\texttt{Marital} & \texttt{Duration} &\texttt{Annual.pct.driven} &\texttt{Pct.drive.mon} \\
\texttt{Insured.sex}&\texttt{Insured.age} &\texttt{Pct.drive.xhrs}  &\texttt{Pct.drive.tue} \\
\texttt{Car.use}&\texttt{Car.age} &\texttt{Pct.drive.rushxx} &\hspace{1cm}. \\
\texttt{Region}&\texttt{Credit.score} & &\hspace{1cm}. \\
\texttt{Territory}&\texttt{Annual.miles.drive} & &\texttt{Pct.drive.sun} \\
\texttt{NB\_Claim}&\texttt{Years.noclaims} & &\texttt{Pct.drive.wkday} \\
 &\texttt{Total.miles.driven} & &\texttt{Pct.drive.wkend} \\
 &\texttt{Avgdays.week} & & \\
 &\texttt{Accel.xxmiles} & & \\
 &\texttt{Brake.xxmiles} & & \\
 &\texttt{Left.turn.intensityxx} & & \\
 &\texttt{Right.turn.intensityxx} & & \\
 &\texttt{AMT\_Claim} & & \\
\hline \hline 
\end{tabular}}
\caption{Data types of all the 52 variables in the synthetic dataset.} \label{tab:TV}
\end{table}

\section{The data generating process} \label{sec:datagen}

The data generation of the synthetic portfolio of 100,000 drivers is a three-stage process using the feedforward neural networks to perform the two simulations and using extended \texttt{SMOTE} to reproduce the feature space.  The first stage is simulating values for the number of claims as multiple binary classifications using feedforward neural networks. The second stage is simulating values for amount of claims as a regression using feedforward neural network with number of claims treated as one of the feature variables. In the final stage, a synthetic portfolio of the space of feature variables is generated applying an extended \texttt{SMOTE} algorithm. The final synthetic data is created by combining the synthetic number of claims, the synthetic amount of claims, and finally, the synthetic portfolio. The resulting data generation is evaluated with a comparison between the synthetic data and the real data when Poisson and gamma regression models are fitted to the respective data. Note that the response variables were generated with extremely complex and nonparametric procedure, so that these comparisons do not necessarily reflect the true nature of the data generation. We also provide other visualization and data summarization to demonstrate the remarkable similar statistics between the two datasets. 

\subsection{The detailed simulation procedures} \label{sub:pro}

Synthetic telematics data is generated based on two feedforward neural network simulations and extended \texttt{SMOTE}. For convenience, we will use notations $\boldsymbol{x}_i\in X=\{X_1,X_2,\cdots,X_{50}\}$, $i=1,2,\cdots, M$, which describe the portfolio having 50 feature variables and $\boldsymbol{x}_i$ is observation (the policy). $Y_1$ is NB\_Claim and $Y_2$ is AMT\_Claim. Superscript $r$ means real data and  $s$ means synthetic data.
	
\subsubsection{The simulation of number of claims} \label{sub:pro1}

To mimic the real telematics data, the first step is to build the simulation generating $Y^s_1$, with four categorical values. It is a multi-class classification problem. However, we converted it into multiple binary class classifications to make each process simple and simultaneously improve the accuracy of simulation.      

\begin{enumerate}
\item Sub-simulation 1: $Z^r_1 = \1{Y^r_1 \geq 1} $. Corresponding instance index is $\{1^{(1)},2^{(1)},\cdots,M^{(1)}\}$. The data is given as the following:\\
 $$\mathcal{D}_1=\{(\boldsymbol{x}^r_{1^{(1)}},z^r_{11^{(1)}}),(\boldsymbol{x}^r_{2^{(1)}},z^r_{12^{(1)}}),\cdots, (\boldsymbol{x}^r_{M^{(1)}},z^r_{1M^{(1)}})\}$$
 \item Sub-simulation 2: $Z^r_2 = \1{Y^r_1 \geq 2|Y^r_1 \geq 1} $. Corresponding instance index is $\{1^{(2)},2^{(2)},\cdots,M^{(2)}\}$. The data is given as the following:\\
 $$\mathcal{D}_2=\{(\boldsymbol{x}^r_{1^{(2)}},z^r_{21^{(2)}}),(\boldsymbol{x}^r_{2^{(2)}},z^r_{2{2^{(2)}}}),\cdots, (\boldsymbol{x}^r_{M^{(2)}},z^r_{2M^{(2)}})\}$$
 \item Sub-simulation 3: $Z^r_3 = \1{Y^r_1 \geq 3|Y^r_1 \geq 2} $. Corresponding instance index is $\{1^{(3)},2^{(3)},\cdots,M^{(3)}\}$. The data is given as the following:\\
 $$\mathcal{D}_3=\{(\boldsymbol{x}^r_{1^{(3)}},z^r_{31^{(3)}}),(\boldsymbol{x}^r_{2^{(3)}},z^r_{3{2^{(3)}}}),\cdots, (\boldsymbol{x}^r_{M^{(3)}},z^r_{3M^{(3)}})\}$$
\end{enumerate}

\noindent Feedforward neural network simulation is learned from each $\mathcal{D}_k$. Hyperparameters are tuned via Gaussian Process (\texttt{GP}) algorithm as detailed in the previous section: the number of hidden layers, the number of nodes for first hidden layer, the number of nodes for the rest of the hidden layers, activation functions, batch size, and the learning rate. The resultant architecture of the network is introduced in Table \ref{tab:arch1}. We set up sigmoid activation function for output layer since this is binary problem; it has the value between 0 and 1. Threshold is 0.5 and cross entropy loss function is used. The weight of the neural network is optimized using the  \texttt{Adam} optimizer. In the \texttt{Adam} optimizer, as input values, we need $\alpha\ (\text{learning rate}), \beta_1,\beta_2$, and $\epsilon$. See Algorithm \ref{alg:adam}. In practice, $\beta_1=0.9,\beta_2=0.999$ and $\epsilon=1e^{-08}$ are commonly used and no further tuning is usually done. Thus, we only tuned the learning rate via \texttt{GP}.
	
\begin{table}[htbp]
\centering
\resizebox{!}{0.975cm}{
\begin{tabular}{l|cccccc} \hline \hline
 Architecture & N.hidden L. & N.nodes\_first hidden L. & N.nodes\_rest hidden L. & Activation & BatchSize & Learning R.\\ \hline
 sub-sim1 & 3 & 353	& 68 & ReLU & 85& 0.000667 \\ \hline
 sub-sim2 & 3 & 473 & 67 & ReLU & 18 & 0.001019 \\ \hline
 sub-sim3 & 2 & 60 & 60 & ReLU & 16 & 0.001922 \\ \hline \hline
\end{tabular}}
\caption{The architecture of the three sub-simulations for number of claims.}\label{tab:arch1}
\end{table}

\begin{figure}[htbp]
\centering
\includegraphics[scale=0.75]{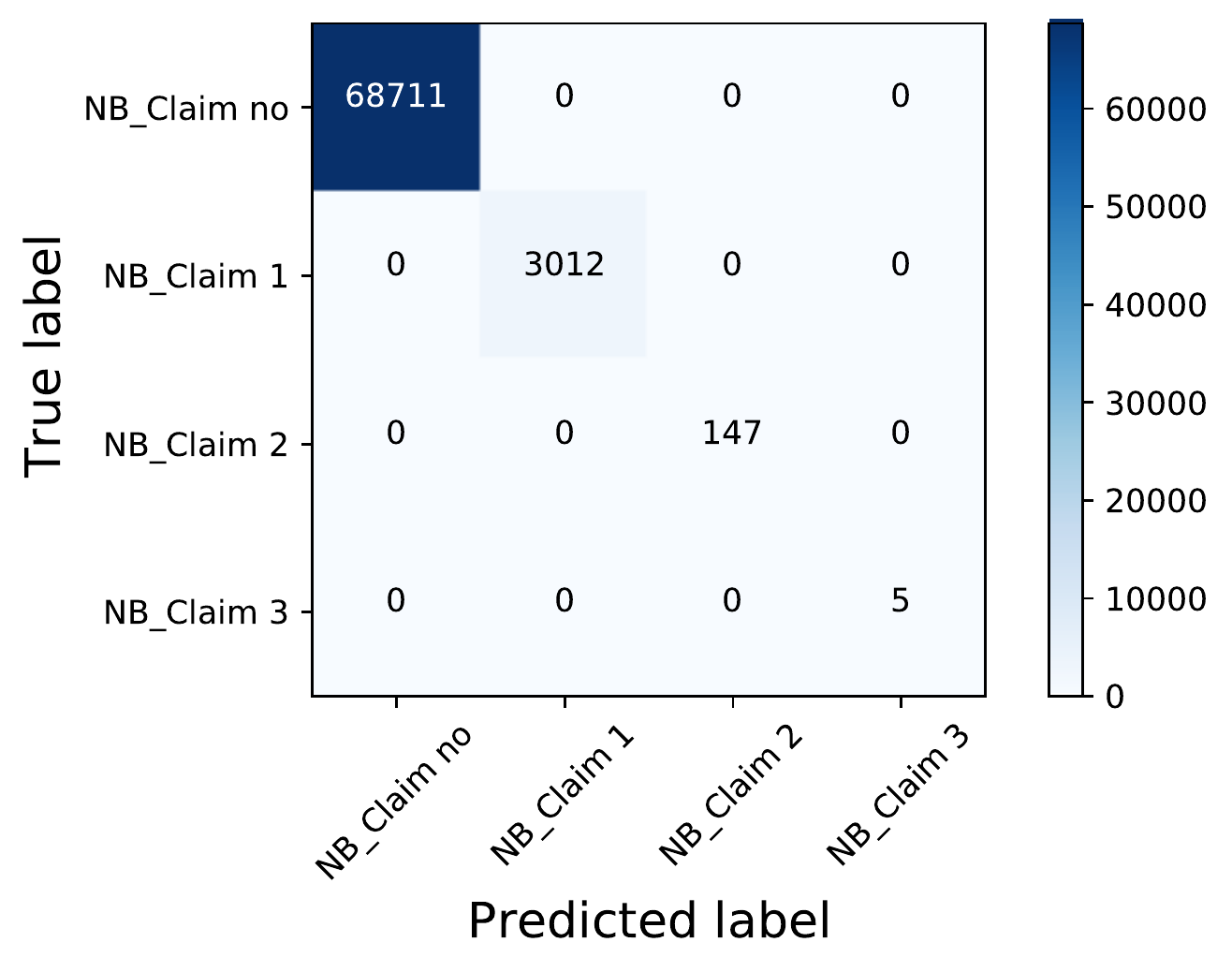}
\caption{Confusion matrix based on the number of claims simulation results.} \label{fig3:conf}
\end{figure}

\noindent The accuracy of the three sub-simulations is shown in Figure \ref{fig3:conf}. When the real portfolio is plugged in, its prediction reveals 100\% coincidence with the real number of claims. This implies that as we plug in realistic portfolio into this combined frequency simulation, we are able to arrive at realistic number of claims.   

After building three sub-simulations, plugging in synthetically generated portfolio, $X^s$ into sub-simulation 1, we get $Z^s_1$. Then we extract $X^s|Z^s_1=1$, plugging it into sub-simulation 2 and get the value, $Z^s_2$. Likewise, plugging in $X^s|Z^s_2=1$ into sub-simulation 3, we obtain the final one, $Z^s_3$. By combining these three results, we finally generate synthetic number of claims, $Y^s_1$.

\begin{algorithm}[htbp]
\KwIn{$\alpha$: Stepsize}
\KwIn{$\beta_1,\beta_2 \in [0,1)$: Exponential decay rates for the moment estimates}
\KwIn{$\epsilon$: }
\KwIn{$f(\theta)$: Stochastic objective function with parameters $\theta$}
\KwIn{$\theta_0$: Initial weights}
	
\KwOut{$\theta_T$: Resulting weights}
 $m_0,v_0,t = 0$: Initialize 1st moment vector, 2nd moment vector, and timestep \;
\For{$t=1, \ldots, T$}{
 $g_t = \nabla_{\theta}f_t(\theta_{t-1}) $: Get gradients w.r.t. stochastic objective at timestep t \;
 $m_t=\beta_1 m_{t-1}+(1-\beta_1)g_t$: Update biased first moment estimate \;
 $v_t=\beta_2 v_{t-1}+(1-\beta_2)g^2_t$: Update biased second raw moment estimate \;
 $\hat{m}_t=m_t/(1-\beta^t_1)$: Compute bias-corrected first moment estimate \;
 $\hat{v}_t=v_t/(1-\beta^t_2)$: Compute bias-corrected second raw moment estimate \;
 $\alpha_t=\alpha \sqrt{1-\beta^t_2}/(1-\beta^t_1)$: Update stepsize \;
 $\theta_t=\theta_{t-1}-\alpha_t m_t / (\sqrt{v_t}+\epsilon)$: Update parameters \;
}
Return $\theta_T$ \;
\caption{Adam (\citet{kingma2014adam})}\label{alg:adam}
\end{algorithm}

\subsubsection{The simulation of aggregated amount of claims} \label{sub:pro2}

We produce the subset of portfolios, which satisfies the condition, $Y^r_1>0$. Corresponding to a new index of the subset is defined as $\{1^{(sev)},2^{(sev)},\cdots,M^{(sev)}\}$. The number and amount of claims are not treated independent to each other but rather, the number of claims $Y^r_1$, is also considered as one of the feature variables. Therefore, we use the following data to train the aggregated amount of claims simulation:
$$
\mathcal{D}_4=\{((\boldsymbol{x}^r_{1^{(sev)}},y^r_{11^{(sev)}}),y^r_{21^{(sev)}}),((\boldsymbol{x}^r_{2^{(sev)}},y^r_{12^{(sev)}}),y^r_{22^{(sev)}}),\cdots, ((\boldsymbol{x}^r_{M^{(sev)}},y^r_{1M^{(sev)}}),y^r_{2M^{(sev)}})\}
$$

$Y^r_2$ is a non-negative continuous value. Thus, in the second simulation, we use \texttt{ReLU} as the activation function and \texttt{MSE} as the loss function. \texttt{Adam} optimizers are used with the hyperparameters selected in the same manner as described in Section \ref{sub:pro1}. These are further described in Table \ref{tab:arch2}.

\begin{table}[htbp]
\centering
\resizebox{!}{0.575cm}{
\begin{tabular}{l|cccccc} \hline \hline
Architecture & N.hidden L. & N.nodes\_first hidden L. & N.nodes\_rest hidden L. & Activation & BatchSize &Learning R.  \\ \hline
 & 6 & 344	& 67 & ReLU & 3 & 0.000526  \\ \hline \hline
\end{tabular}}
\caption{The architecture of simulation for the aggregated amount of claims.}\label{tab:arch2}
\end{table}

Figure \ref{fig3:qq} reveals the resulting performance of the claims simulation. Prediction errors are highly centered around zero and most of dots are on the line of QQ plot for predicted and real claim amount. This sufficiently proves that the simulation can imitate the real amount of claim with synthetic portfolio based on the number of claims simulation introduced in Section \ref{sub:pro1}.  

\begin{figure}[H]
\centering
\includegraphics[scale=0.57]{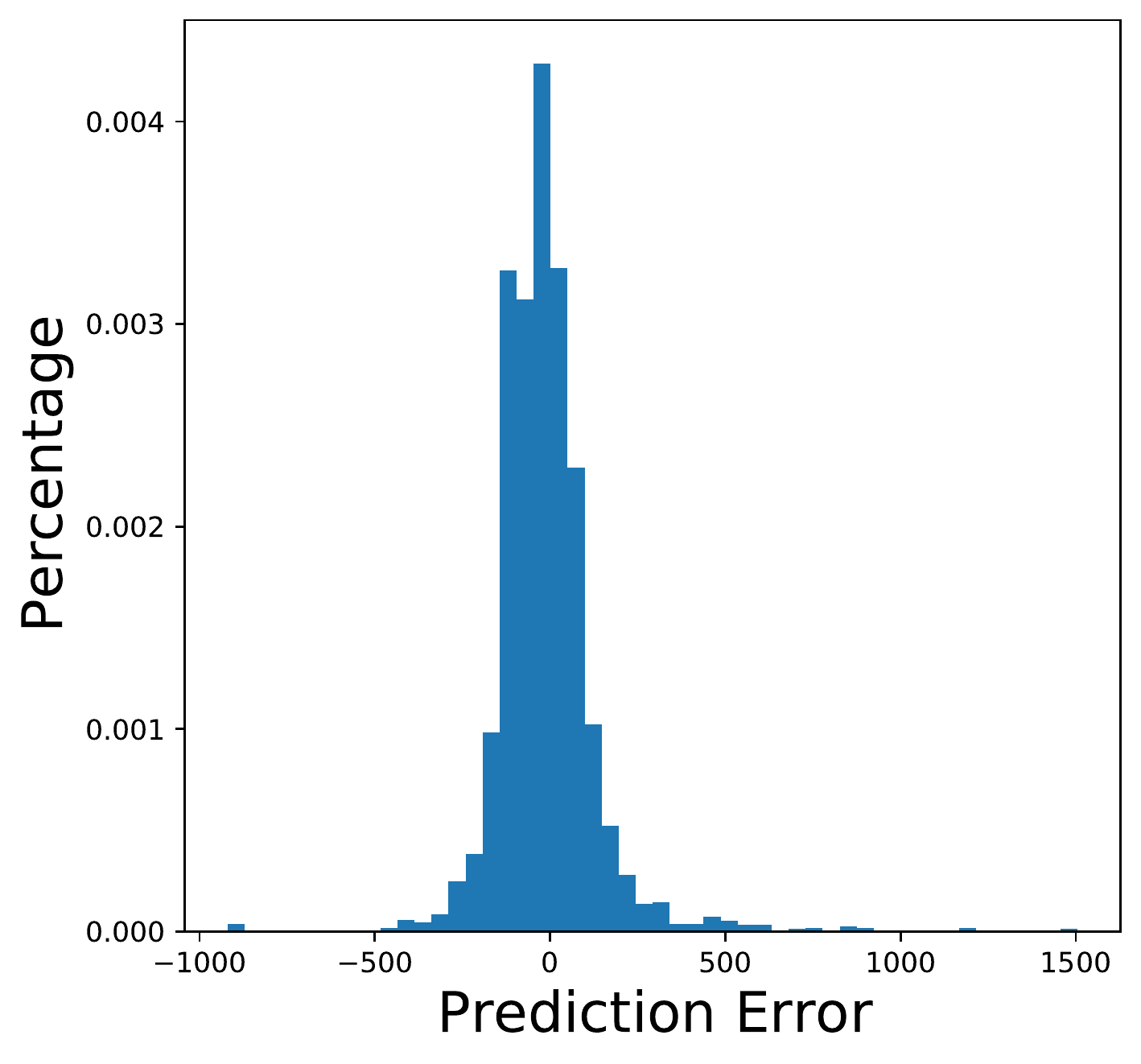}
\includegraphics[scale=0.57]{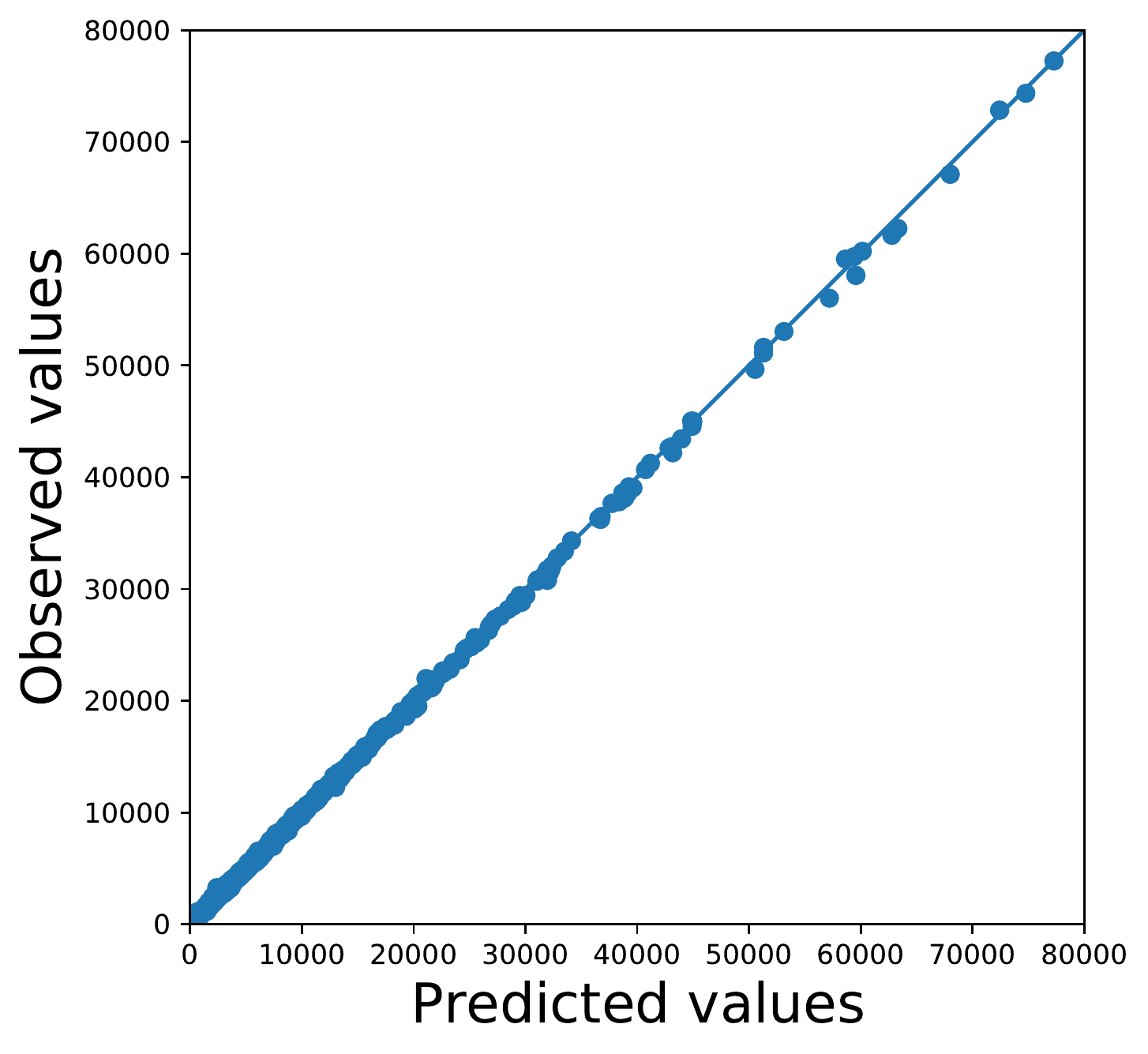}
\caption{Assessing the accuracy of the simulation of aggregated amount of claims.} \label{fig4:aggregate}
\end{figure}

To generate $Y^s_2$, we use $Y^s_1$ obtained from Section \ref{sub:pro1} and we extract the subset of synthetic portfolio with the condition, $Y^s_1 > 0$. This subset of synthetic portfolio and corresponding $Y^s_1$ are the input of the simulation to get $Y^s_2$.

\subsubsection{Synthetic portfolio generation} \label{sub:pro3}

\begin{figure}[H]
\centering
\includegraphics[scale=0.75]{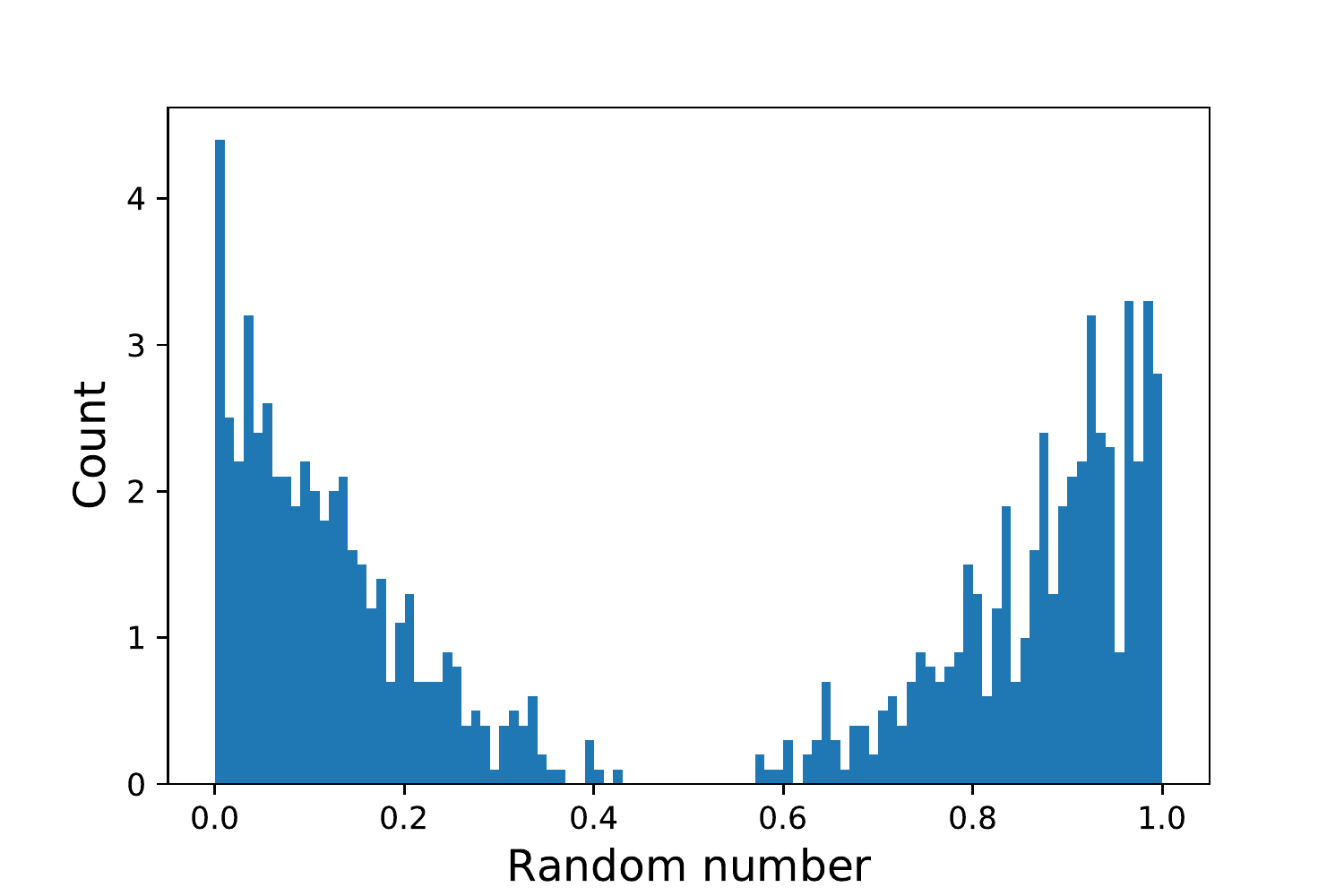}
\caption{1000 random numbers drawn from the $U$-shape distribution} \label{fig4:random}
\end{figure}

As described in Section \ref{sub:smote}, we propose extended version of \texttt{SMOTE} to generate the final synthetic portfolio, $X^s$. Extended \texttt{SMOTE} is primarily different from the original \texttt{SMOTE} in just a single step: the interpolation step. The detailed procedure is the following: for each feature vector (observation, $x^r_i$), the distance between $x^r_i$ and the other feature vectors in $X^r$ are computed based on the Euclidean distance and one-nearest neighbor is obtained. Difference between $x^r_i$ and this neighbor is multiplied by a random number drawn from the $U$-shape distribution as shown in Figure \ref{fig4:random}. Adding the random number to the $x^r_i$, we create a synthetic feature vector, $x^s_i$. 100,000 synthetic observations are generated, which consisted of the synthetic portfolio, $X^s$. After applying the extended \texttt{SMOTE}, the following considerations had also been reflected in the synthetic portfolio generation.  

\begin{itemize}
\setlength\itemsep{0.0em}
\item Integer features are rounded up;
\item For categorical features, only \texttt{Car.use} are multi-class. \texttt{Car.use} is converted by one-hot coding before applying extended \texttt{SMOTE} so that every categorical feature variable has the value 0 or 1. After the generation, they are rounded up;
\item For compositional features, \texttt{Pct.drive.sun} and \texttt{Pct.drive.wkend} are not involved in the generation process but are calculated by `1 $-$ the rest of  related features.' 	
\end{itemize}

\subsection{Comparison: Poisson and gamma regression} \label{sec:sim}

Combining every outputs ($X^s, Y^s_1, Y^s_2$) obtained from Section \ref{sub:pro}, the data with telematics features is thereby complete. Any statistical or machine learning algorithms can now be performed on this completed synthetic datafile. To further compare the quality of the reconstruction of the real dataset to produce the synthetic datafile, one simple approach is to compare the resulting outputs when a Poisson regression model is calibrated on the number of claims (frequency) and a gamma regression model is calibrated on the amount of claims (severity), using the respective real dataset and the synthetic datafile. Both models are relatively standard benchmark models in practice. To be more specific, we fitted both Poisson and gamma regression models to the real and synthetic data to predict the number of claims ($\frac{\text{NB\_Claim}}{\text{Duration}}$) and the average amount of claims ($\frac{\text{AMT\_Claim}}{\text{NB\_Claim}}|\text{{\scriptsize NB\_Claim $>$ 0}}$). A net premium can be calculated by taking the product of the number of claims and the average amount of claims. The purpose of this exercise is not to evaluate the quality of the models nor the relative importance of the feature variables, but rather to compare the resulting outputs between the two datasets. The training models are based on all the feature variables  in the absence of variable selection.

\begin{figure}[H]
\centering
\includegraphics[width=17cm, height=4cm]{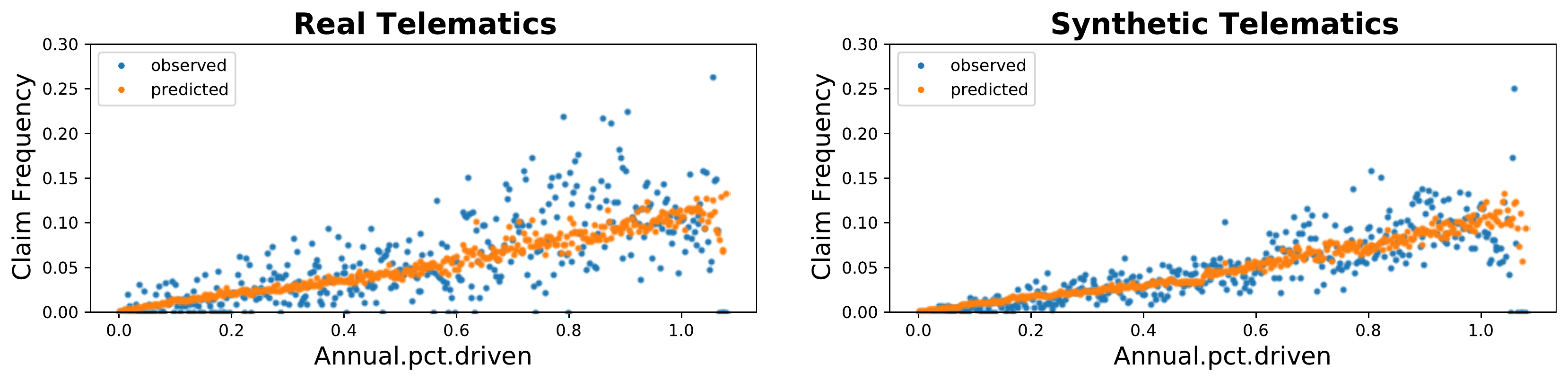} \\
\includegraphics[width=17cm, height=4cm]{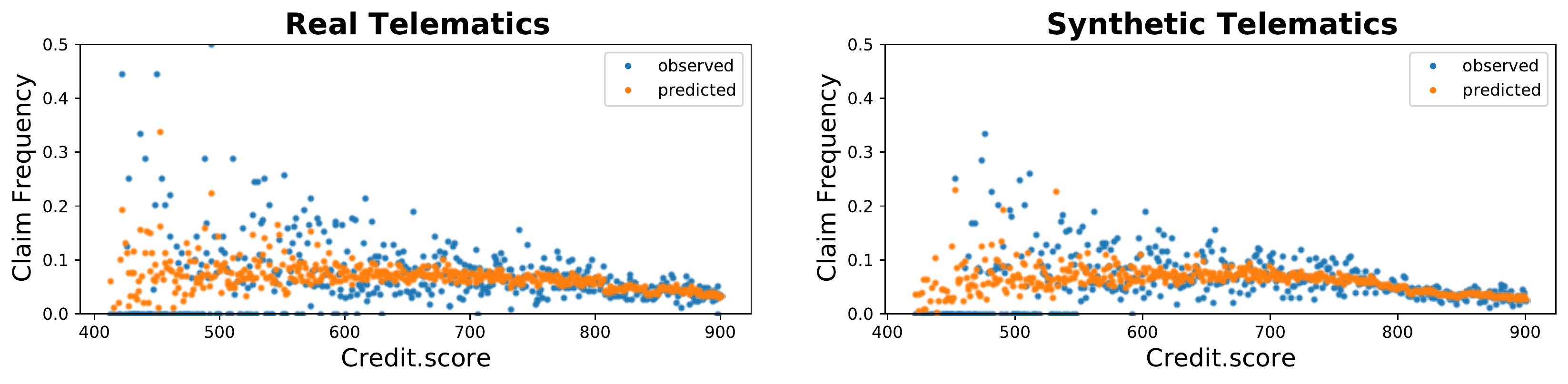} \\
\includegraphics[width=17cm, height=4cm]{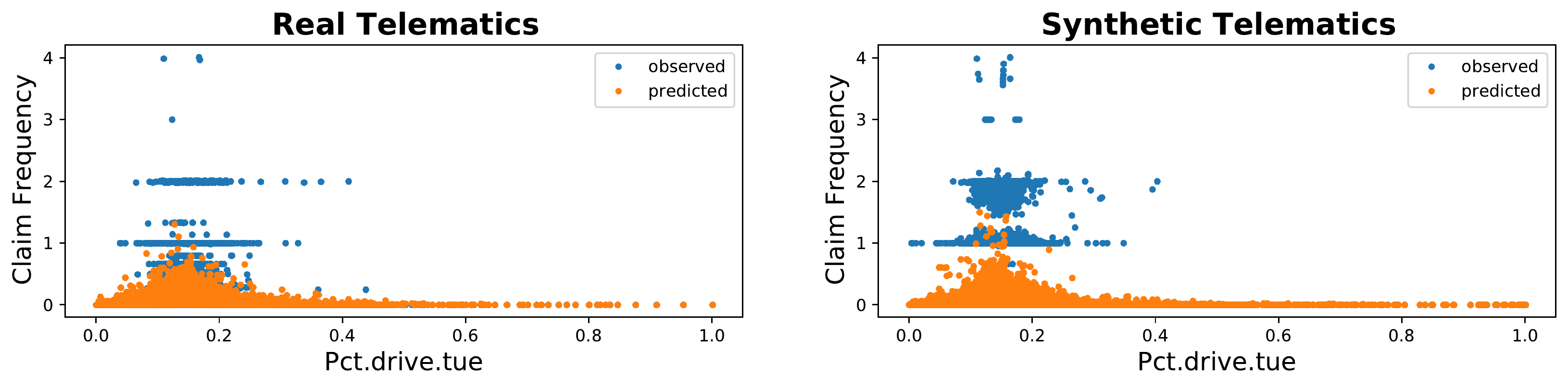} \\
\caption{Average claim frequency using real (left) and synthetic (right) datasets.} \label{fig1:freq}
\end{figure}

Figure \ref{fig1:freq} describes the average claim frequency between the real telematics on the left side and the synthetic telematics on the right side. For simplicity, we only provide the behavior of the claim frequency for three feature variables: \texttt{Annual.pct.drive}, \texttt{Credit.score}, and \texttt{Pct.drive.tue}. For both datasets, we see that observed values are colored blue and the predicted values are colored orange. As we expected, the distributions of the average claim frequency, the pattern of blue and orange, for these feature variables considered here have very similar patterns between the real and the synthetic datasets.

\begin{figure}[H]
\centering
\includegraphics[width=17cm, height=8cm]{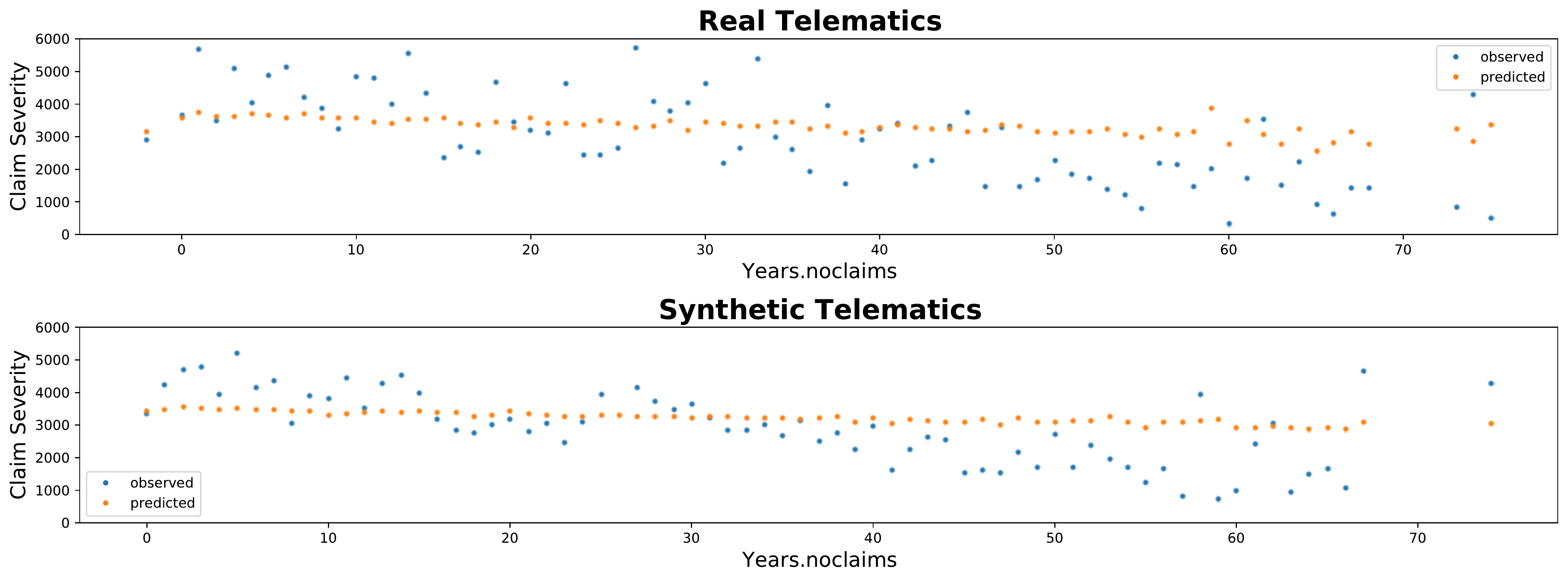} \\
\includegraphics[width=17cm, height=8cm]{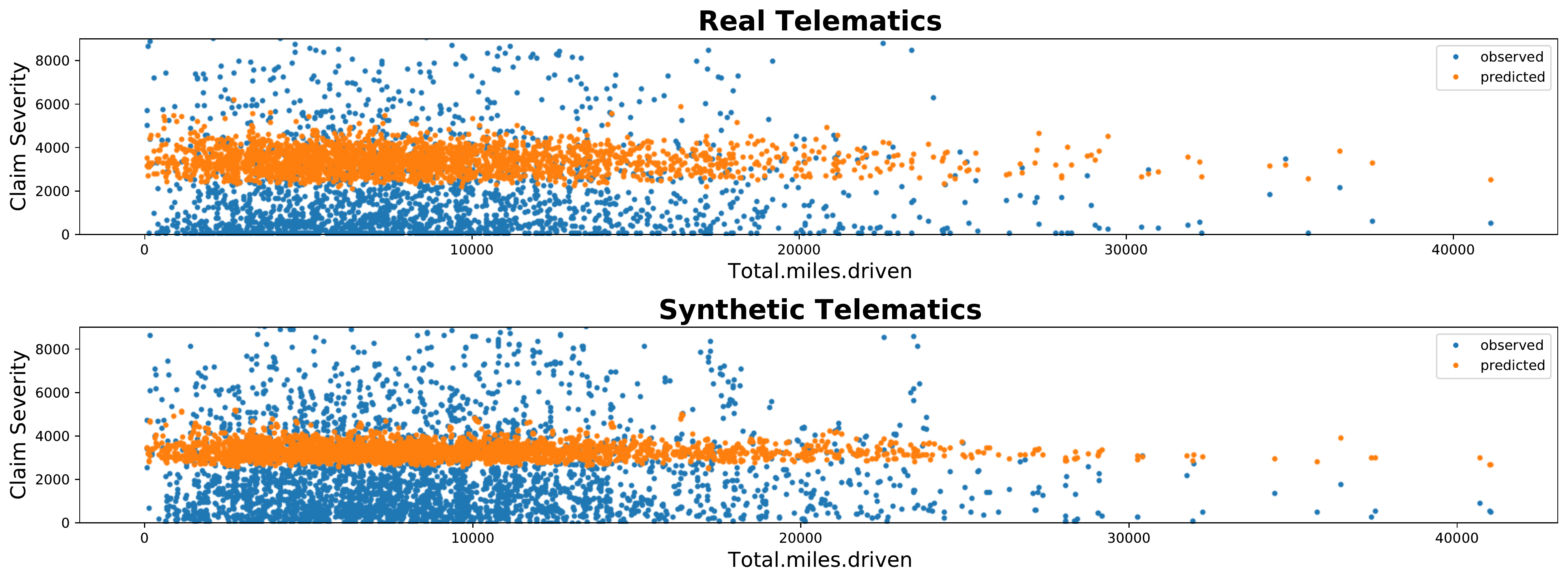} \\
\caption{Average severity using real (1st \& 3rd) and synthetic (2nd \& 4th) datasets.} \label{fig2:sev}
\end{figure}

As similarly done for frequency, Figure \ref{fig2:sev} depicts the average claim severity between the real telematics and the synthetic telematics. For our purpose, we examine these comparisons based on two feature variables: \texttt{Yrs.noclaims} and \texttt{Total.miles.driven}. Both these feature variables do not seem to produce much variation in the predicted values: this may explain that these are relatively less important predictor variables for claims severity. However, this may also be explained by the fact that we do not necessarily have an exceptionally good model here for prediction. However, this is not the purpose of this exercise.

Still from both Figure \ref{fig1:freq} and \ref{fig2:sev}, there are some information we can draw. First, the patterns of blue dots are similar between the real and synthetic data for every feature variable considered here. Even though we do not include the graphs of other features, for all features, they show similar dispersion. Included features are the one considered as importance variables on classification model introduced in \citet{so2020cost}. This seems to suggest that real and synthetic data have similar frequency and feature distributions for all variables, which implies that the synthetic datafile is behaving as realistic as the real data. In conclusion, it mimics the real dataset exceptionally well. Second, the patterns of orange dots are also similar between the real and synthetic data. In more details, predicted frequency (Figure \ref{fig1:freq}) and severity (Figure \ref{fig2:sev}) from model tuned based on real data have similar dispersion with those from model tuned on synthetic data. This suggests results obtained by synthetic data might have little difference from results obtained by real data and we can use synthetic data to train statistical model instead of real data.

\begin{figure}[H]
\centering
\includegraphics[scale=0.55]{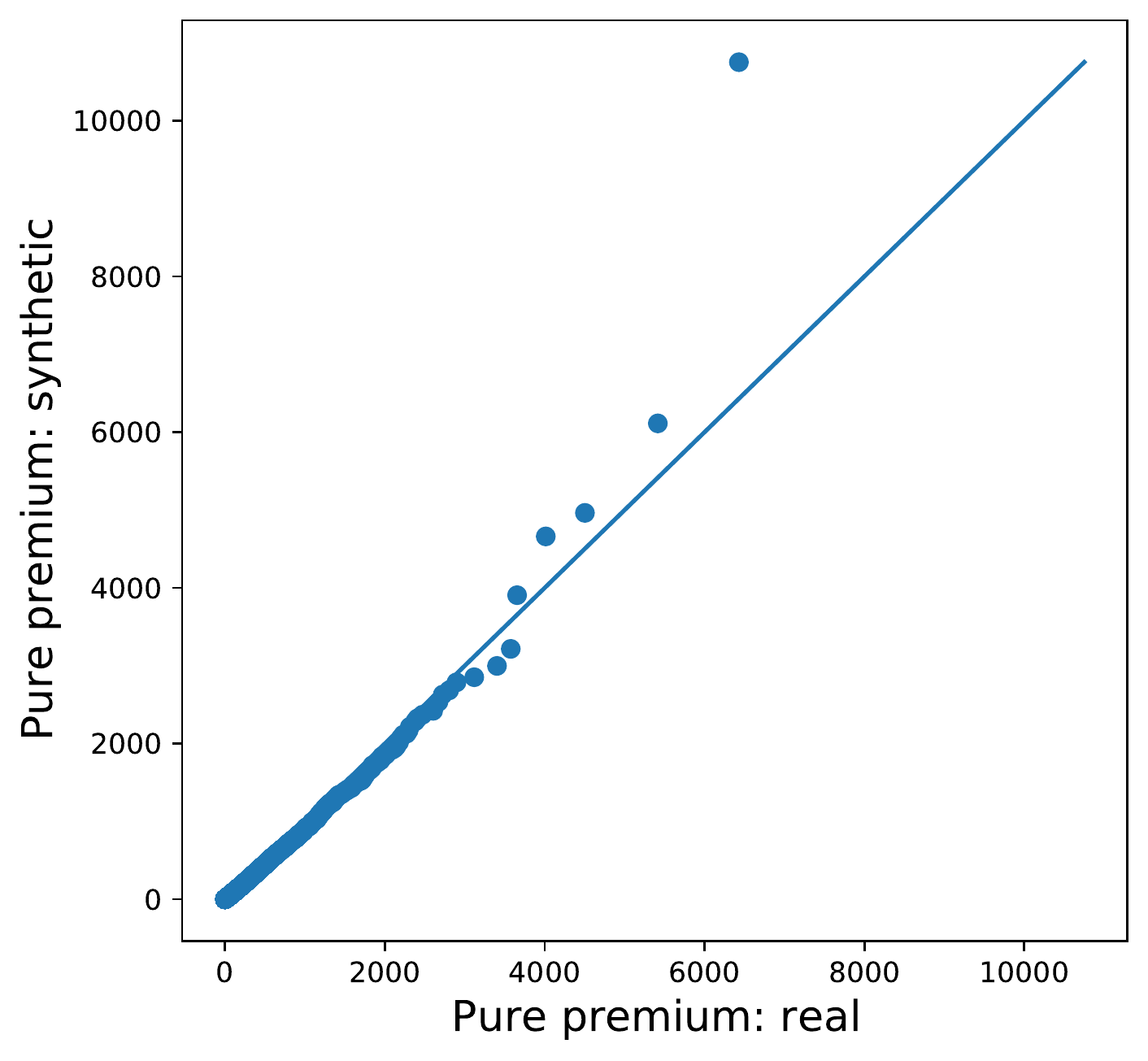}
\caption{QQ-plot of predicted pure premium: real and synthetic data} \label{fig3:qq}
\end{figure}

These conclusions are further supported by Figure \ref{fig3:qq}, which shows quantile-quantile (QQ) plot of the predicted pure premium between the real data and the synthetic data. We do, however, observe that we tend to overestimate the pure premium for the synthetic datafile for high quantiles. This may be a result of the randomness produced throughout the data generation process. This is not, by any means, an alarming concern.

\section{Concluding remarks} \label{sec:conclude}

It has been discussed that there is a perceived positive social effect to vehicle telematics: it encourages careful driving behavior. Indeed, UBI programs can have many potential benefits to insurers, consumers, and the society, in general. Insurers are permitted to put a price tag that links more directly related to habits of insured drivers. As a consequence, this helps insurance companies increase the predictability of their profit margin and provides customers the opportunity for more affordable premiums. On the other hand, consumers may be able to control the level of premium costs by maintaining safer driving habits or if at all possible, by reducing the frequency of driving. Furthermore, UBI may benefit the society because with safer driving and fewer drivers on the road, this may reduce the frequency of accidents, traffic congestion, and car emissions. In order to get the optimal benefits of UBI to both insurers and their policyholders, it becomes subsequently crucial to identify the more significant telematics variables that truly affects the occurrence of car accidents. These perceived positive benefits motivated us to provide the research community a synthetic datafile, which has the intricacies and characteristics of a real data, that may be used to examine, construct, build, and test better predictive models that can immediately be put into practice. For additional details of benefits of UBI, see \citet{husnjak2015UBI}.

In summary, this paper describes the generating process used to produce a synthetic datafile of driver telematics that has largely been based and emulated from a similar real insurance dataset. The final synthetic dataset produced has 100,000 policies that included observations about driver's claims experience, together with associated classical risk variables and telematics-related variables. One primary motivation for such production is to encourage the research community to develop innovative and more powerful predictive models; this synthetic datafile can be used to train and test such predictive models so that we can provide better techniques that can be used in practice to assess UBI products. As alluded throughout this paper, the data generation is best described as a three-stage process using feedforward neural networks to simulate the number and aggregated amount of claims and later applying extended \texttt{SMOTE} algorithm to finalize the portfolio in its entirety. The resulting data generation is evaluated by a comparison between the synthetic data and the real data when Poisson and gamma regression models are fitted to the respective data. Data summarization and visualization of these resulting fitted models between the two datasets produce remarkable similar statistics and patterns.  We are hopeful that researchers interested in obtaining driver telematics datasets to calibrate statistical models or machine learning algorithms will find the output of this research helpful for their purpose. We encourage the research community to build better predictive models and test these models with our synthetic datafile.

\medskip

\subsection*{Appendix. Graphical display of distributions of selected variables between synthetic and real datasets} \label{appa-alg}

The figures in this appendix provide summarization and visualization of selected variables in the datasets. These figures provide suggestions of how remarkably similar the distributions of the two datasets, an indication how good our procedure generated the synthetic datasets. Note that we can only provide distribution summaries in order to preserve confidentiality of the original data used in the generation. The figures are self-explanatory.

\begin{figure}[htbp]
	\centering
	\includegraphics[scale=0.4]{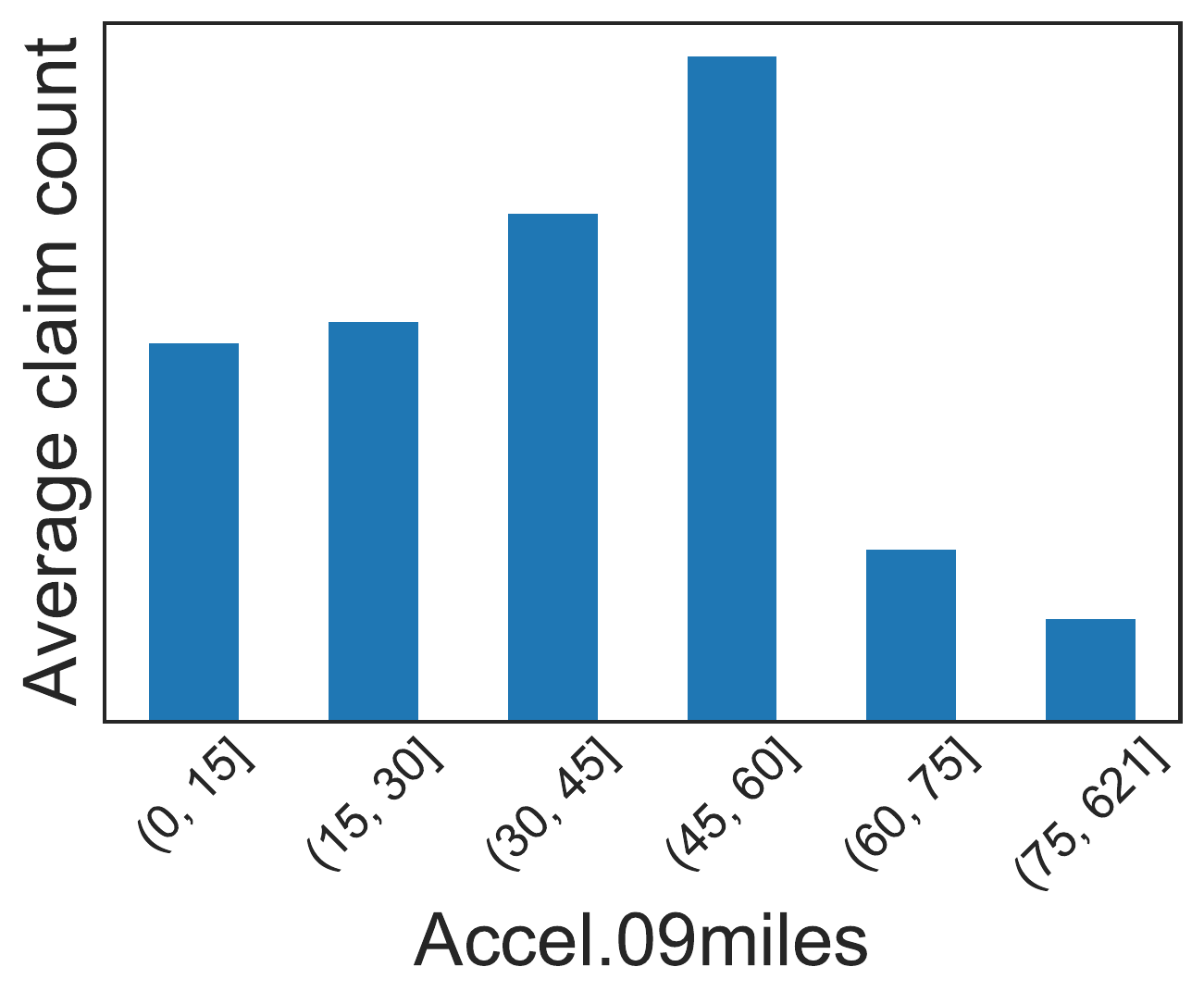} 
	\includegraphics[scale=0.4]{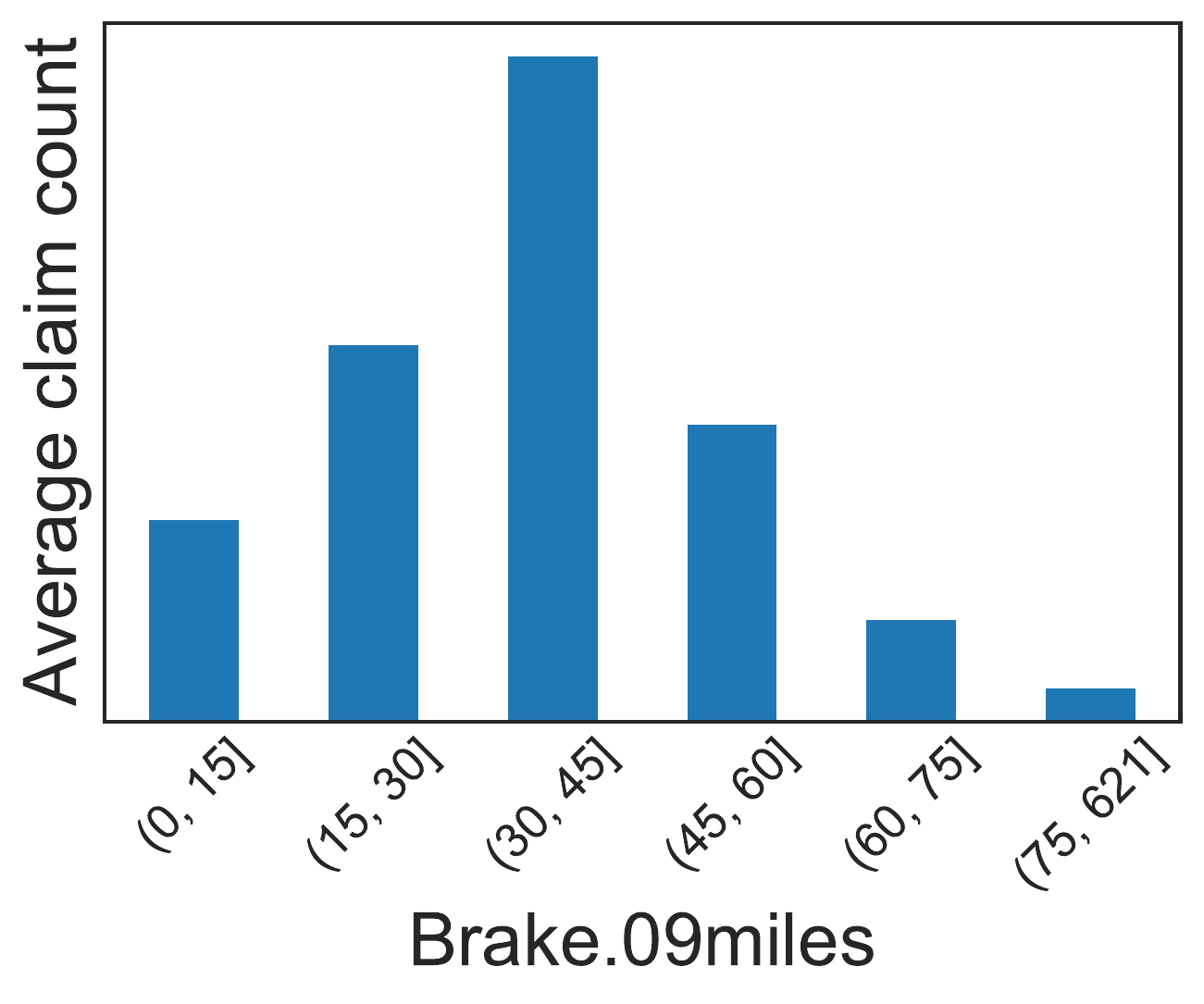}  
	\includegraphics[scale=0.4]{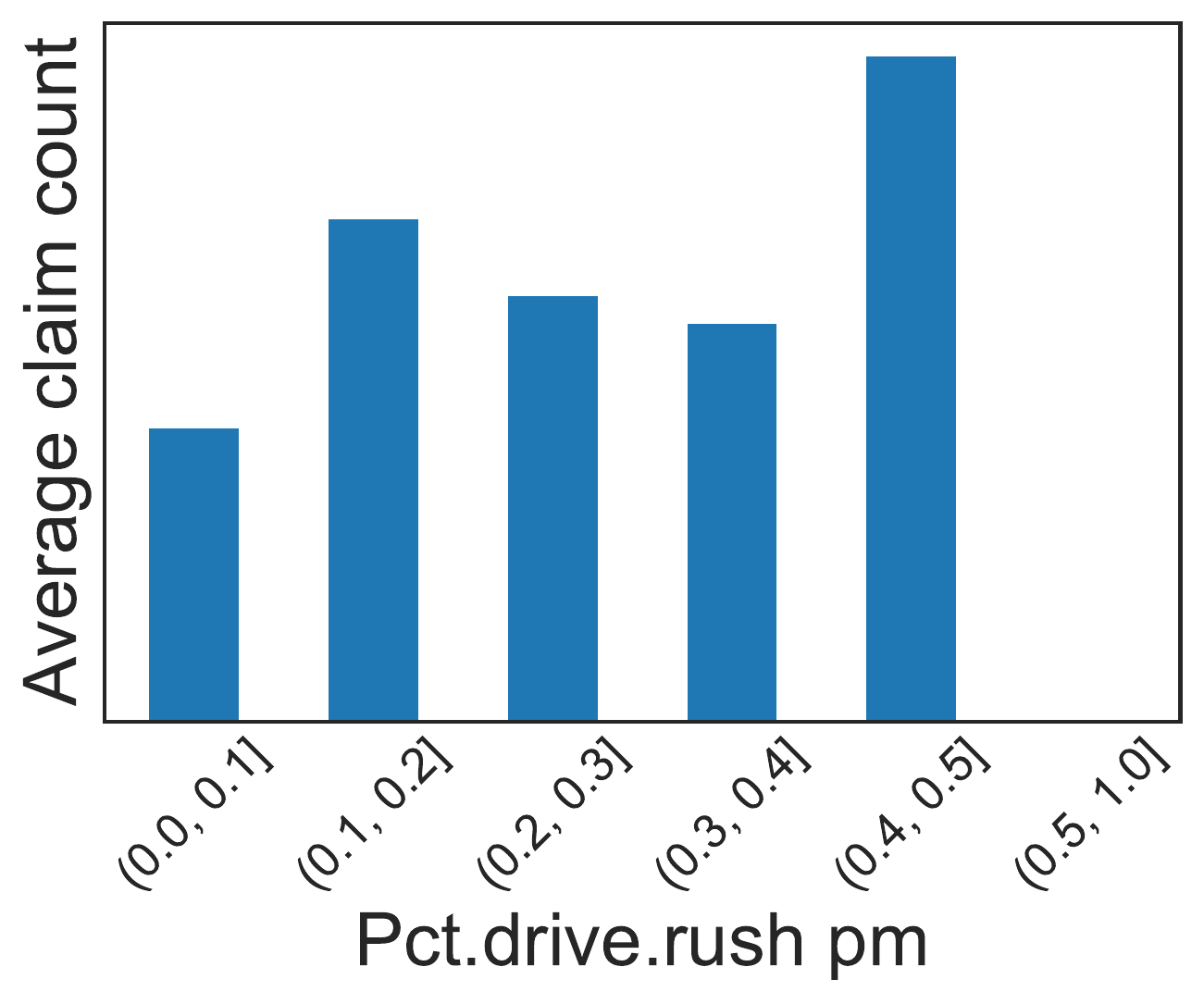} 
	\includegraphics[scale=0.4]{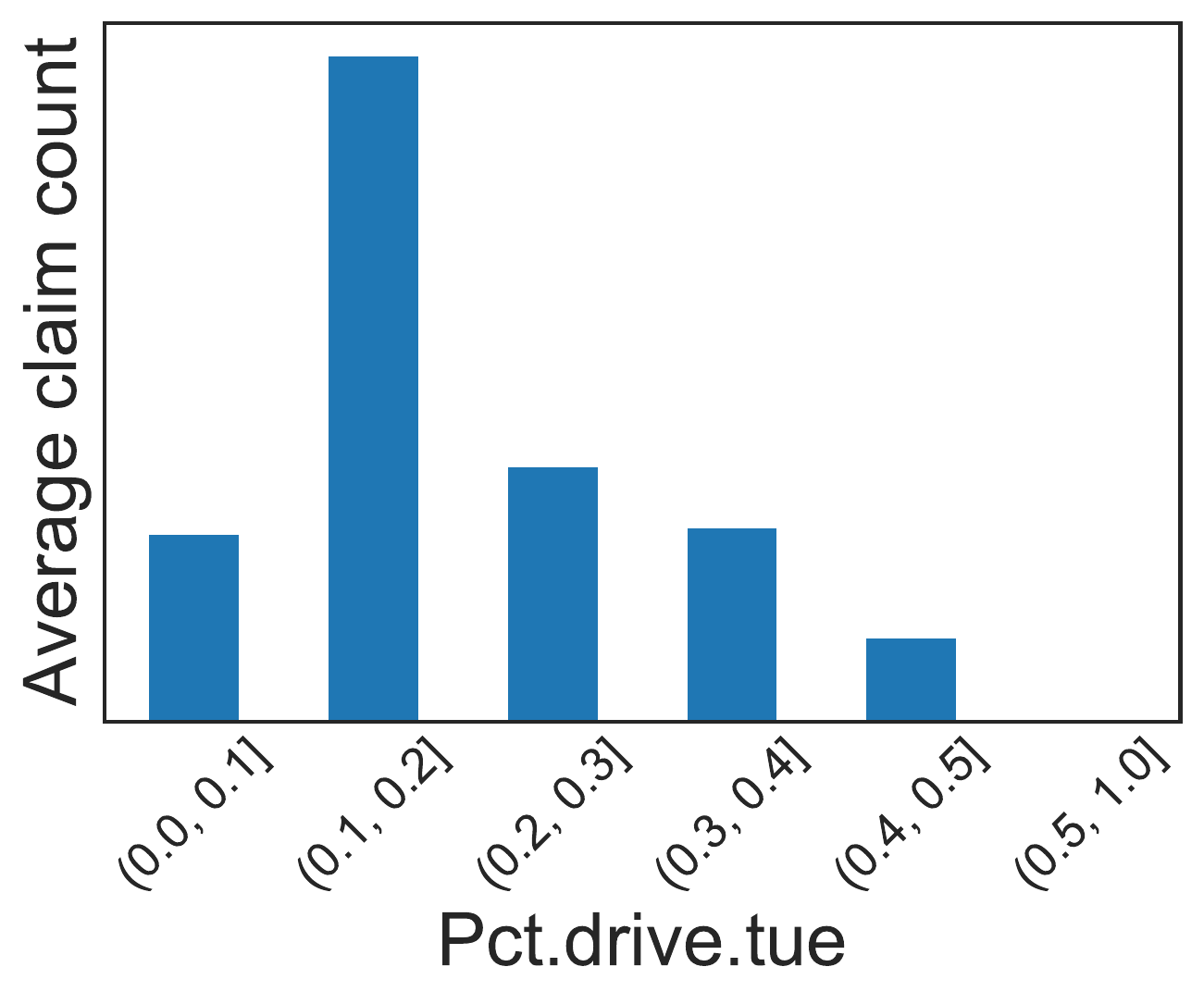} 
	\includegraphics[scale=0.4]{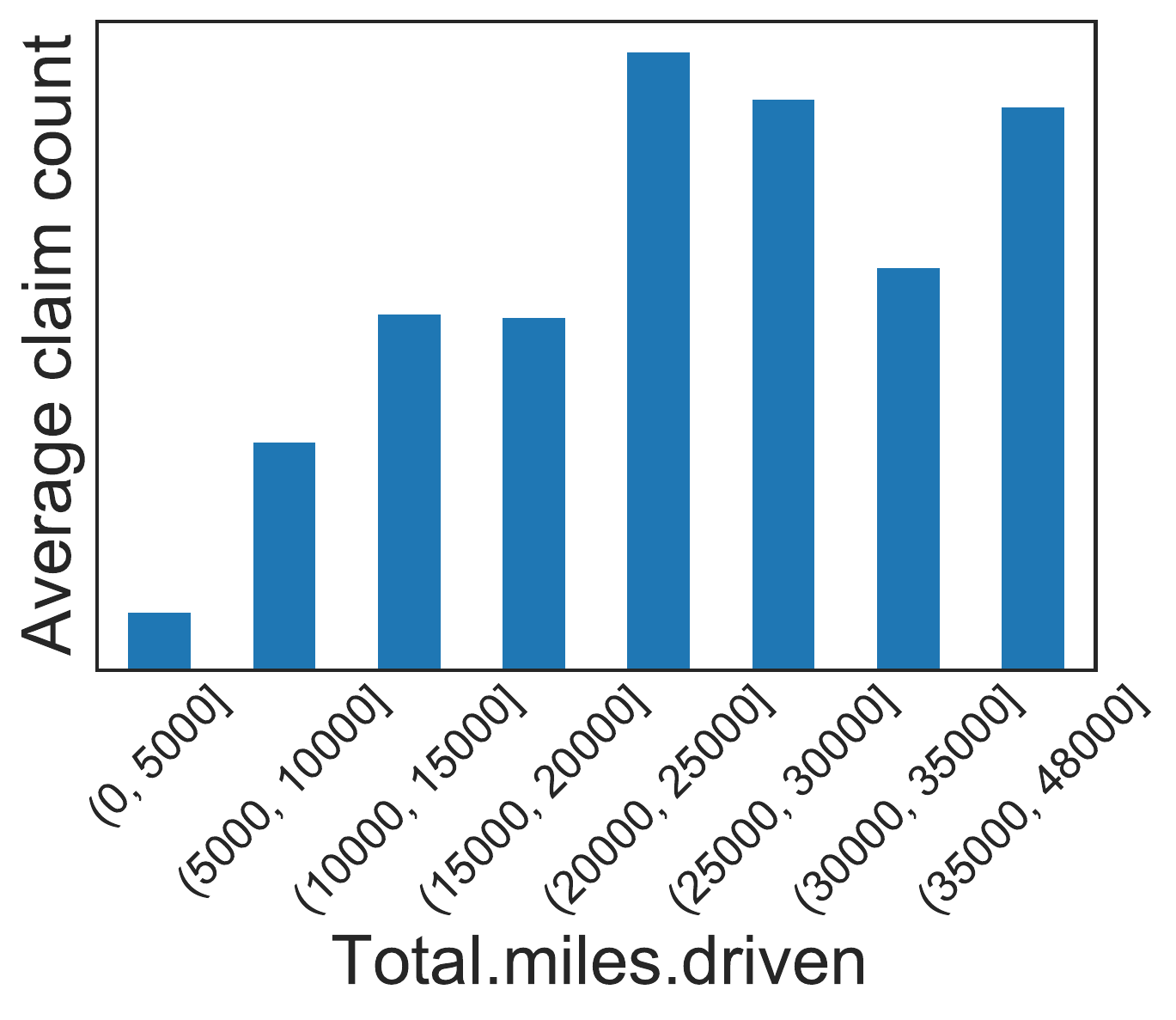} 
	\includegraphics[scale=0.4]{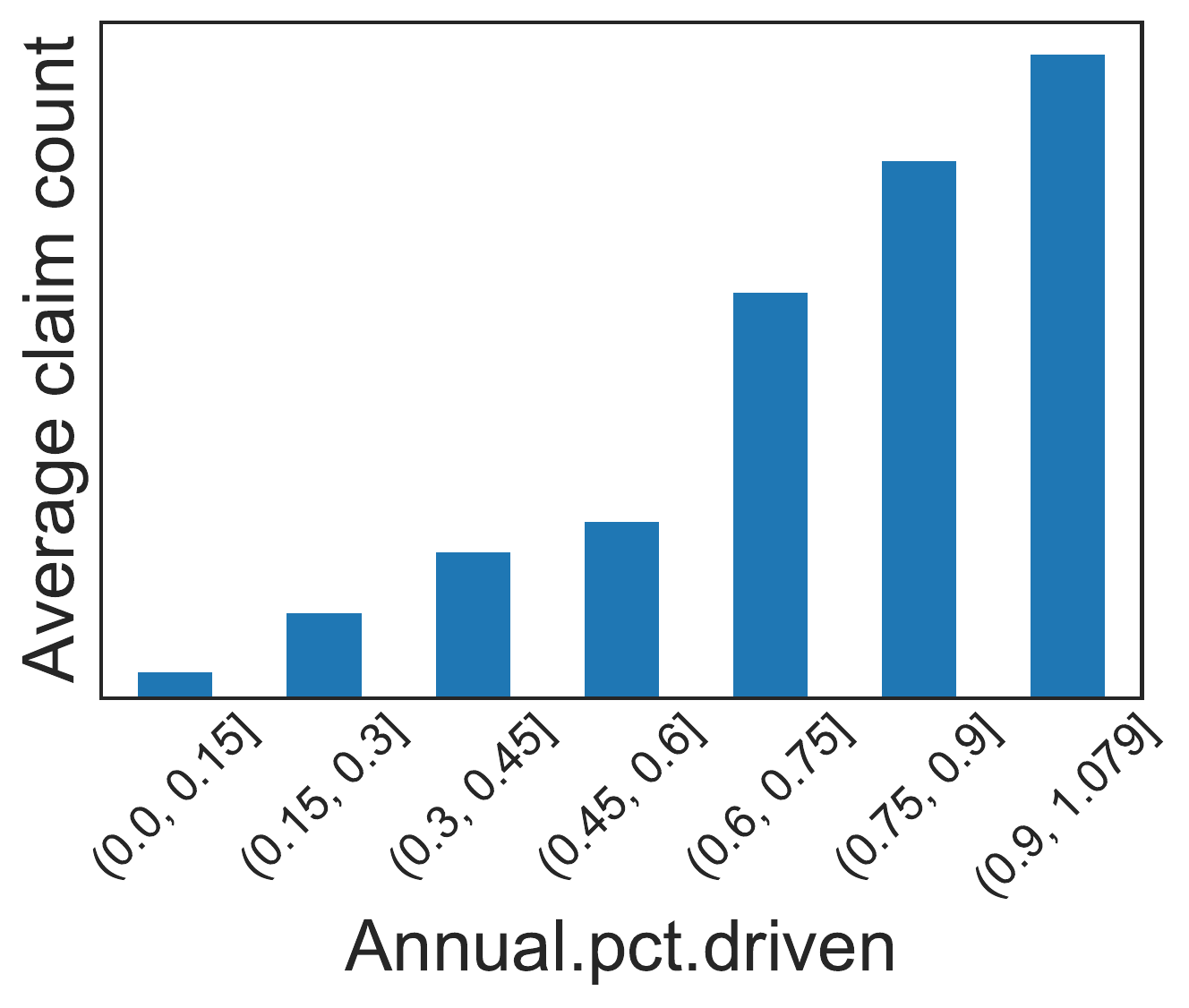}	
	\caption{Synthetic data: Distribution of average number of claims for six telematics-related features.} \label{fig9:dist_tel_syn}
\end{figure}

\begin{figure}[htbp]
	\centering
	\includegraphics[scale=0.4]{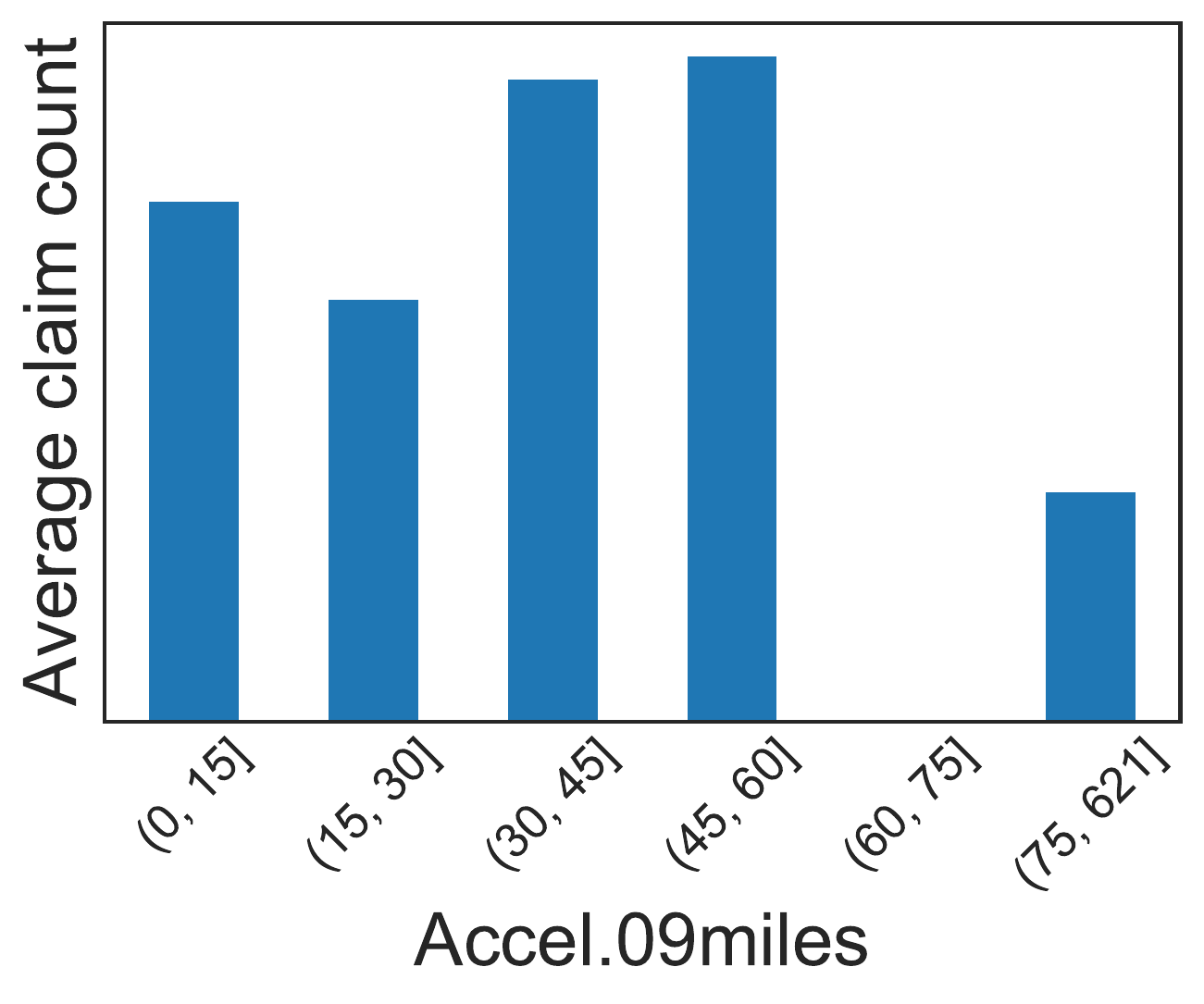} 
	\includegraphics[scale=0.4]{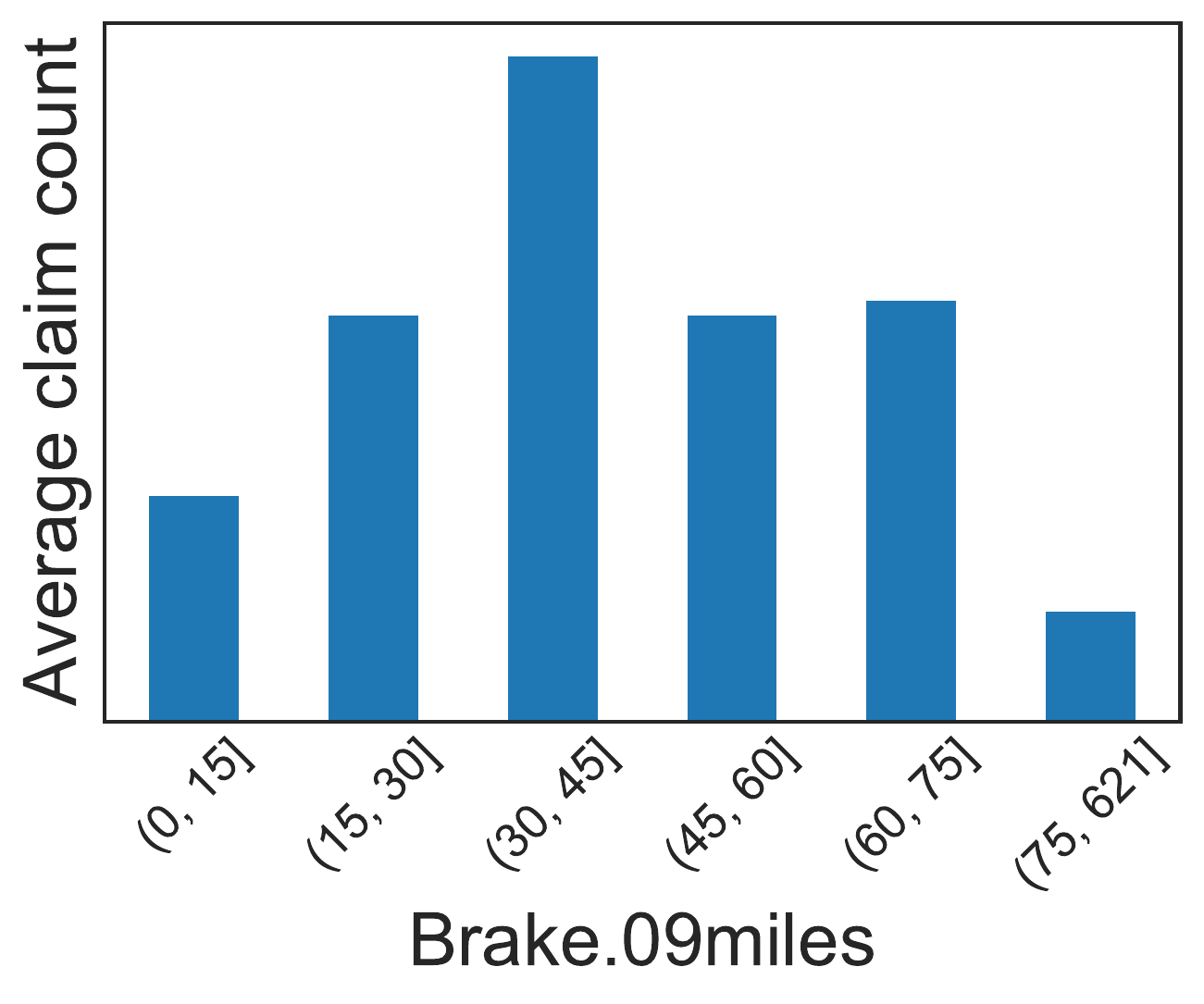} 
	\includegraphics[scale=0.4]{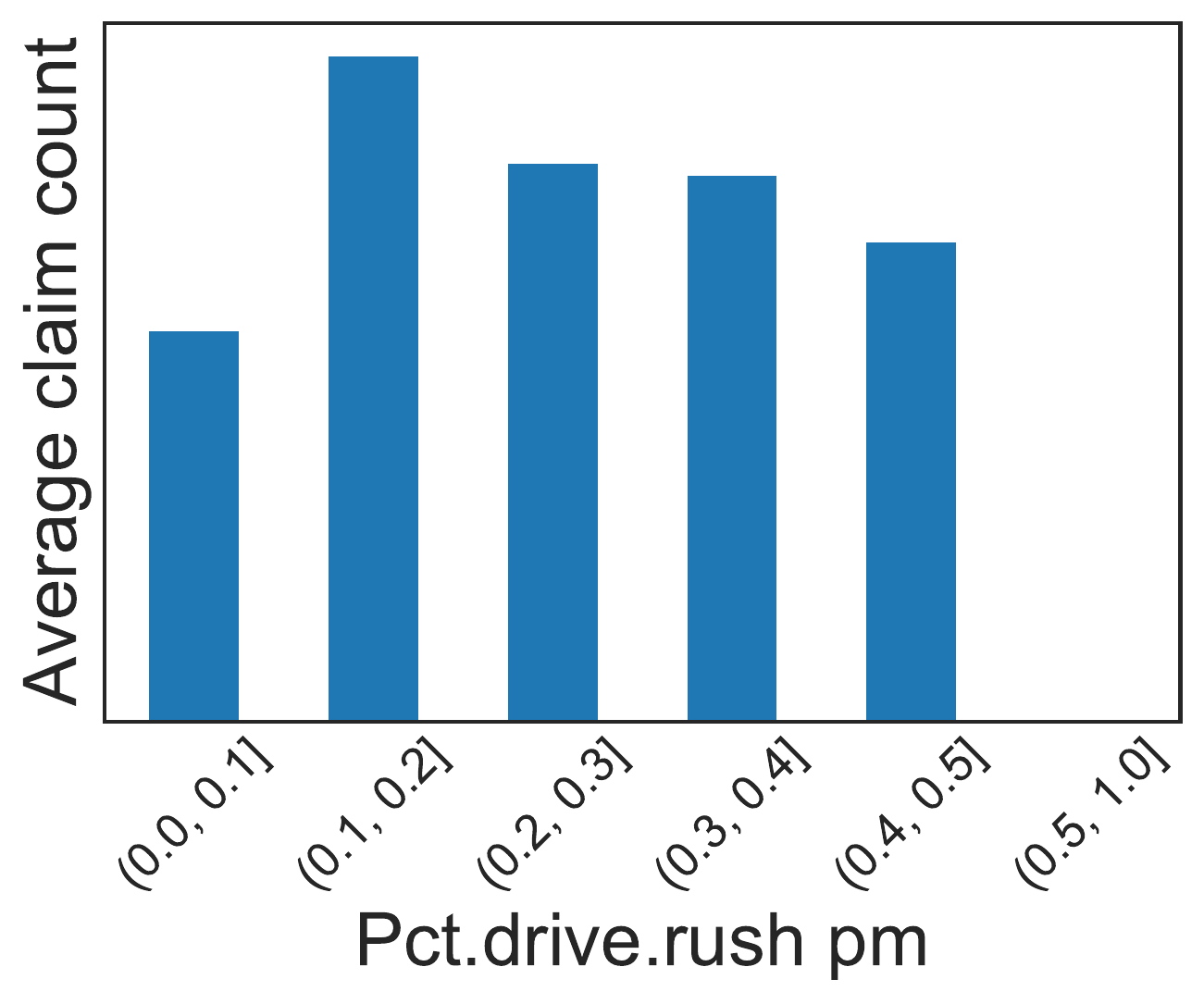} 
	\includegraphics[scale=0.4]{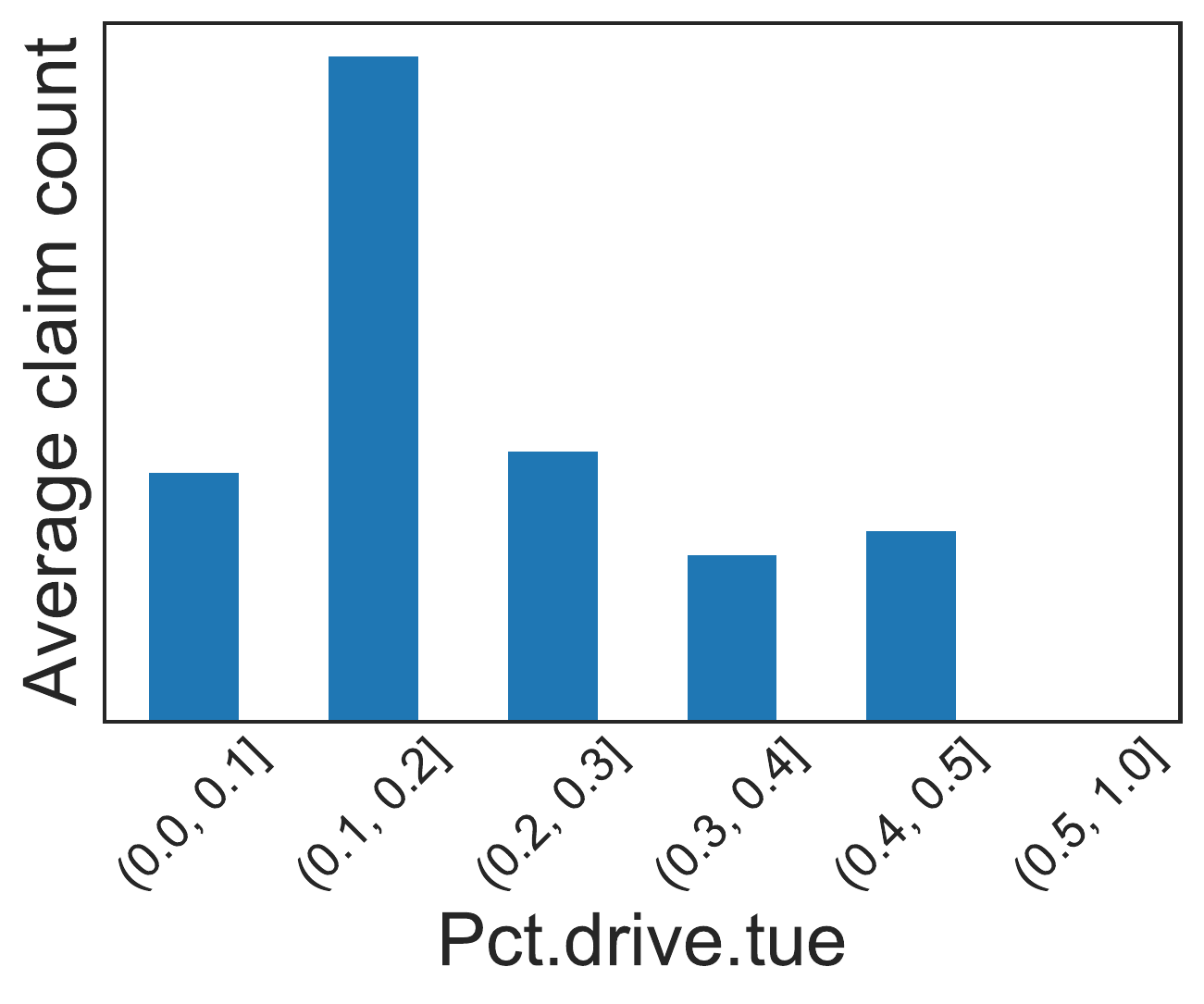} 
	\includegraphics[scale=0.4]{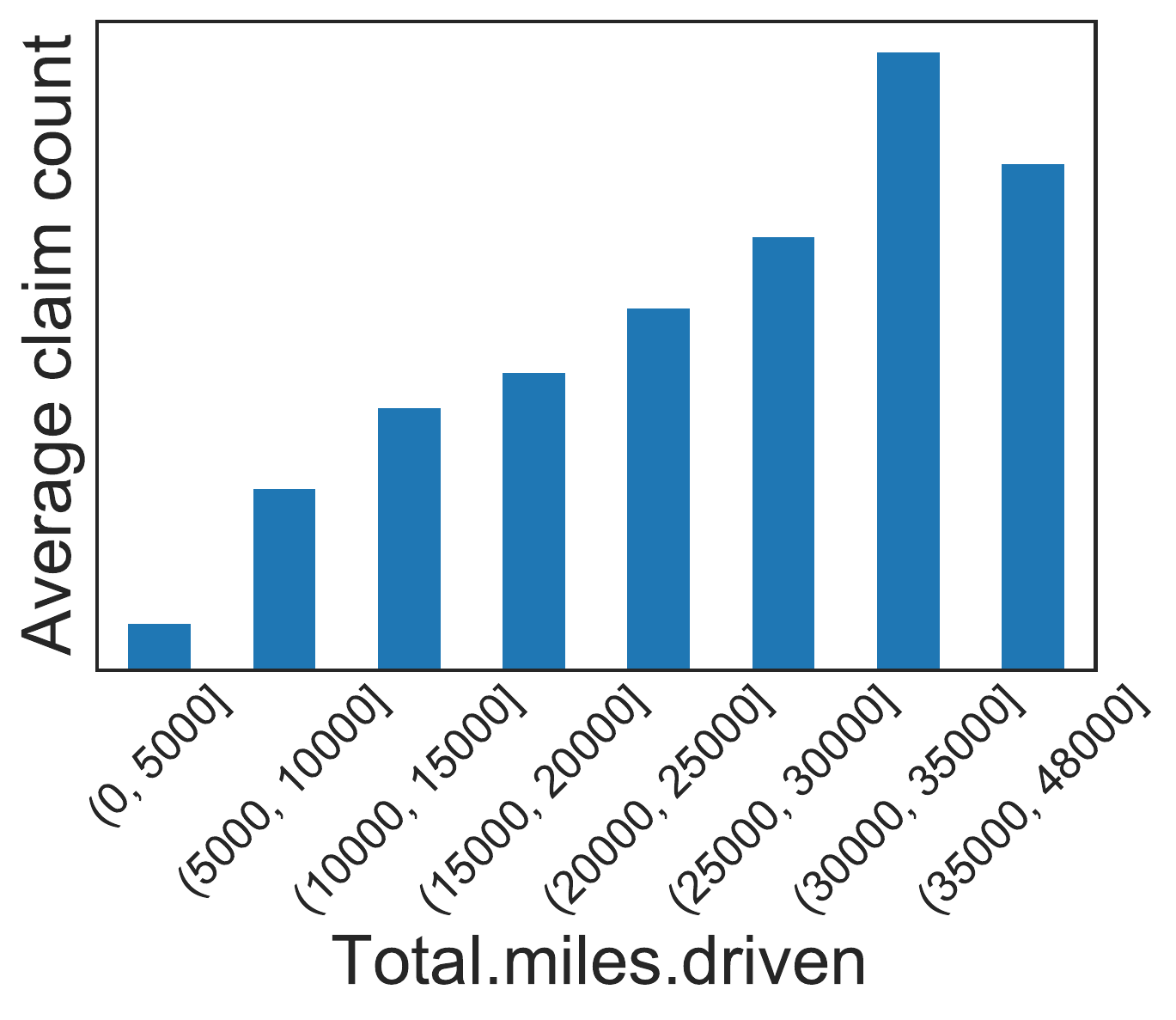} 
	\includegraphics[scale=0.4]{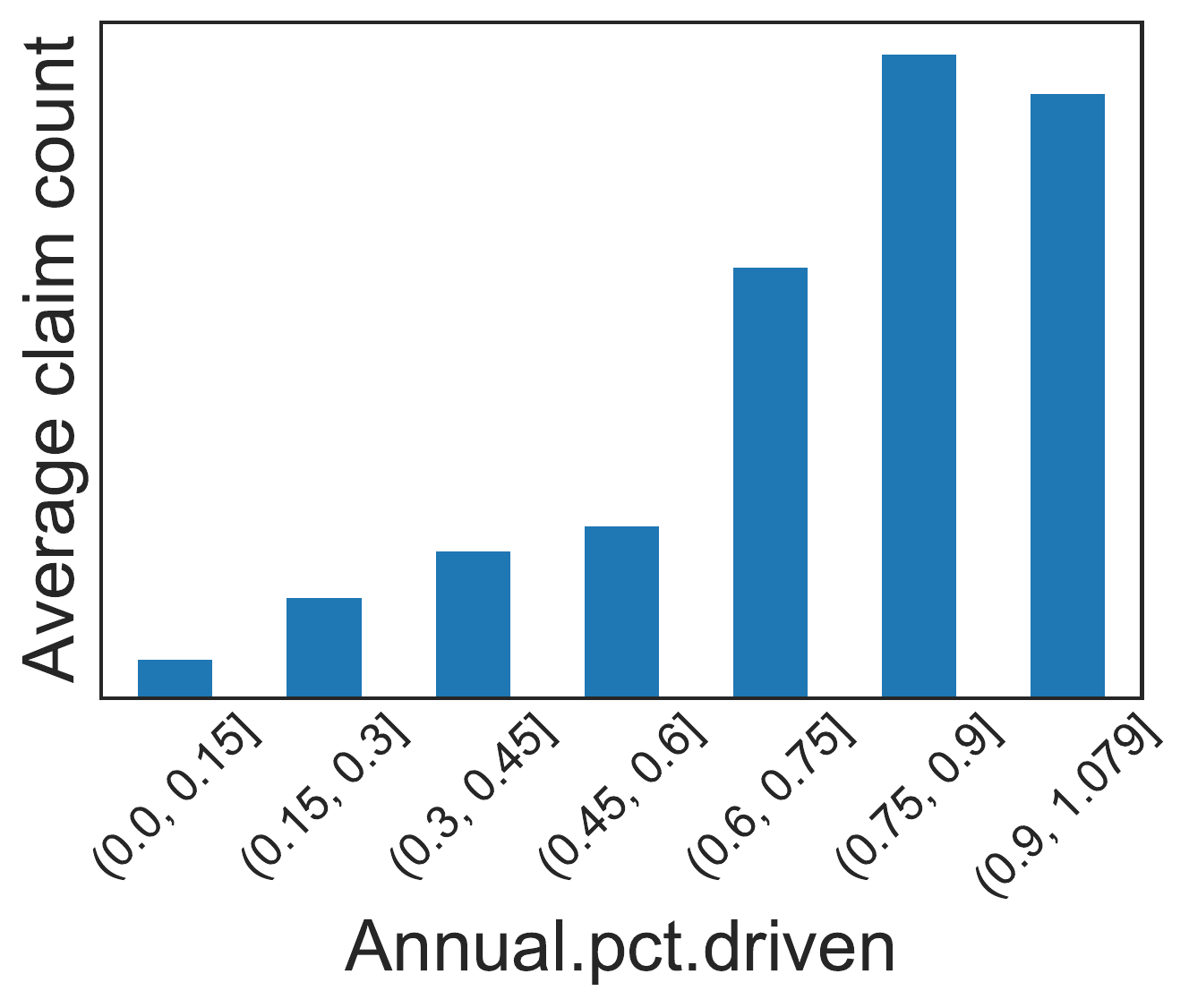} 
	\caption{Real data: Distribution of average number of claims for six telematics-related features.} \label{fig10:dist_tel_real}
\end{figure}

\begin{figure}[htbp]
	\centering
	\includegraphics[scale=0.3]{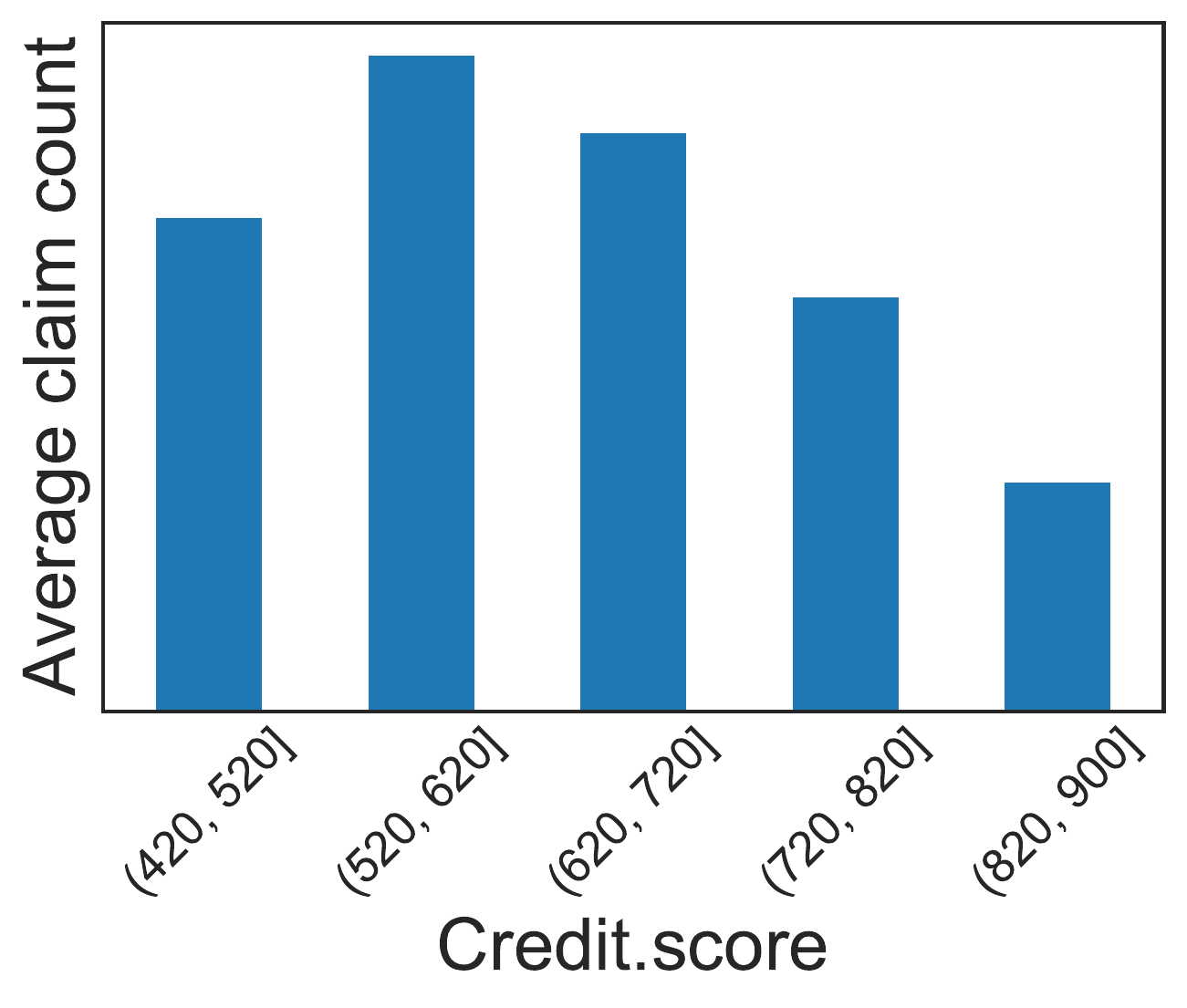} 
	\includegraphics[scale=0.3]{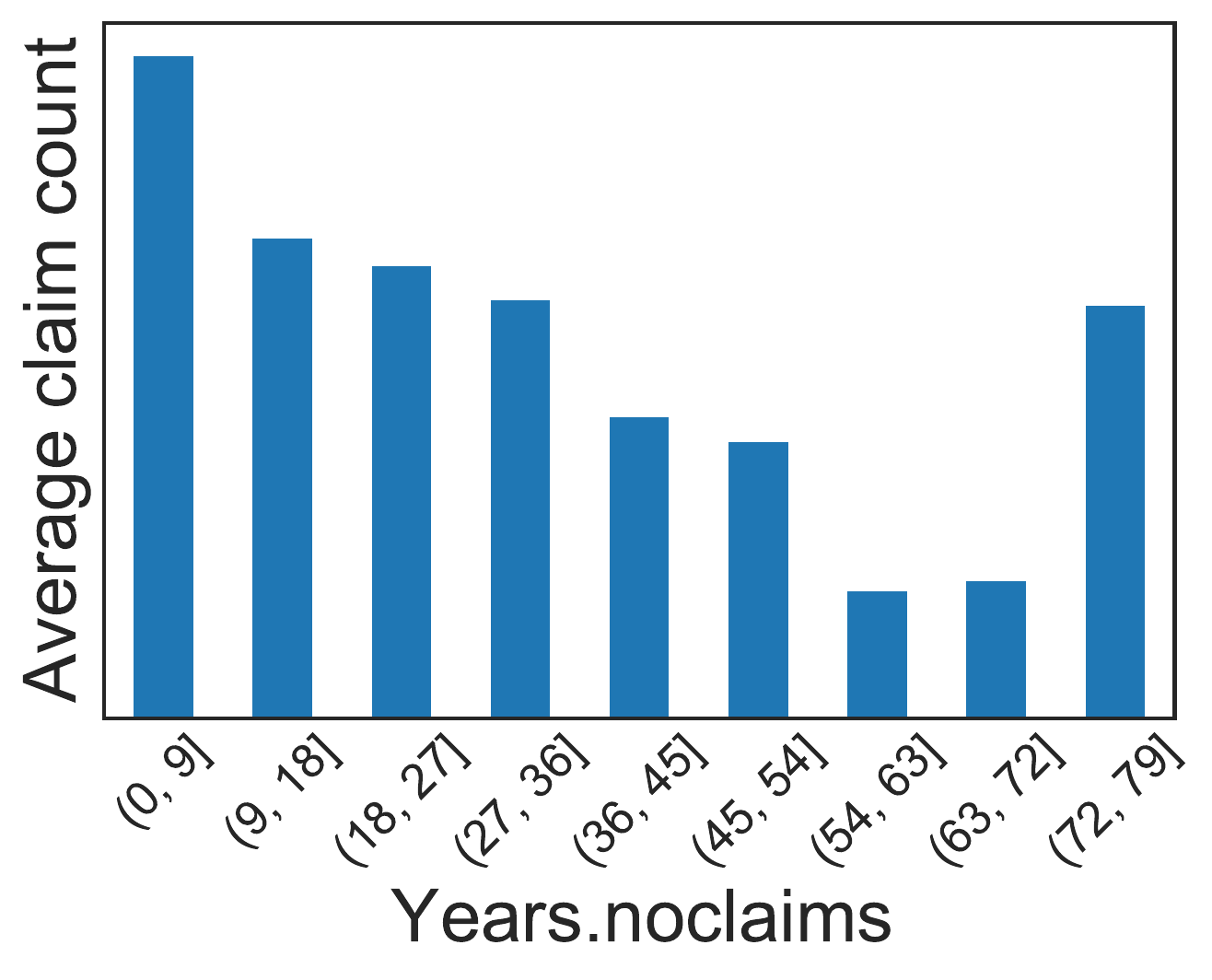} 
	\includegraphics[scale=0.3]{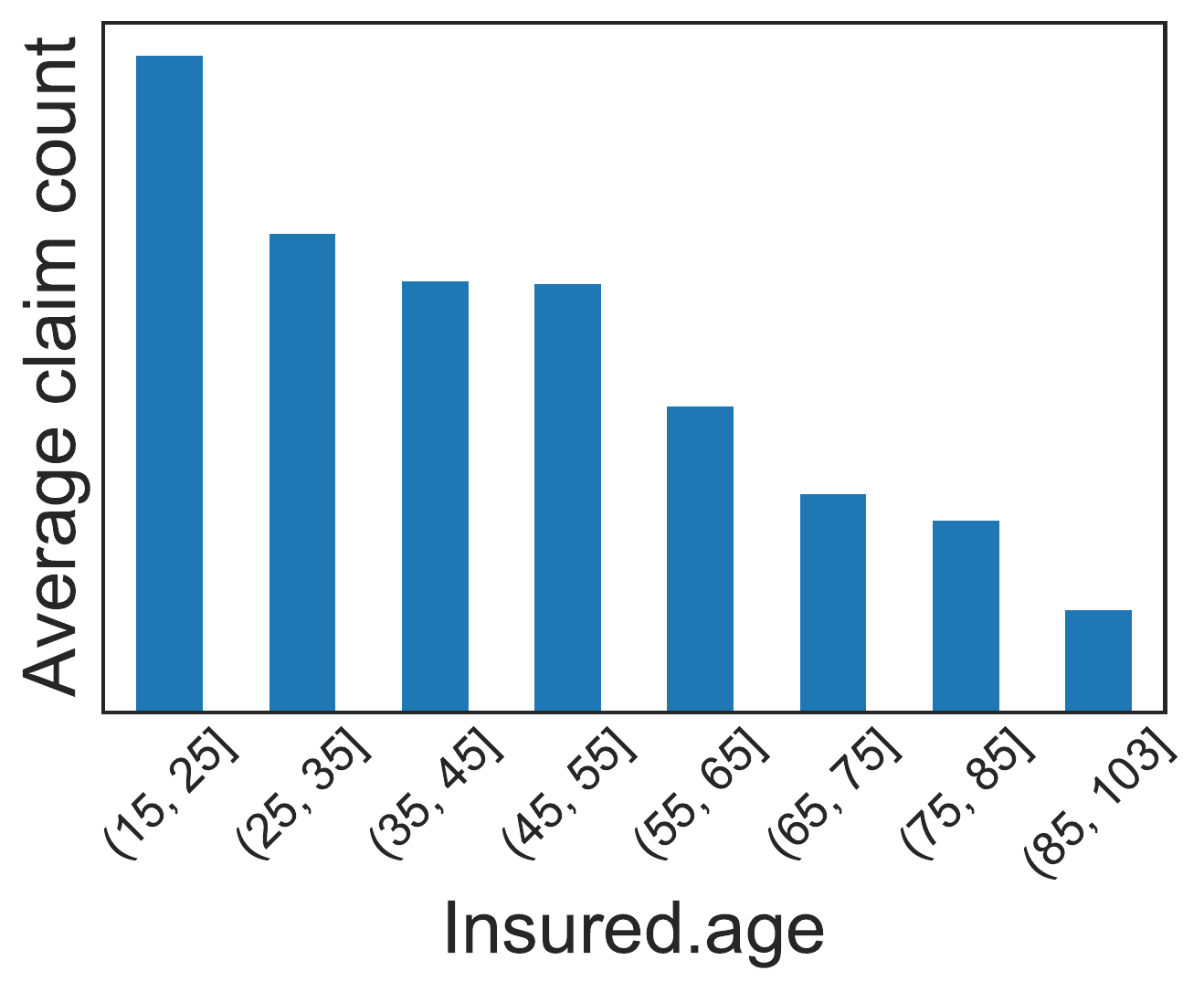} 
	\includegraphics[scale=0.3]{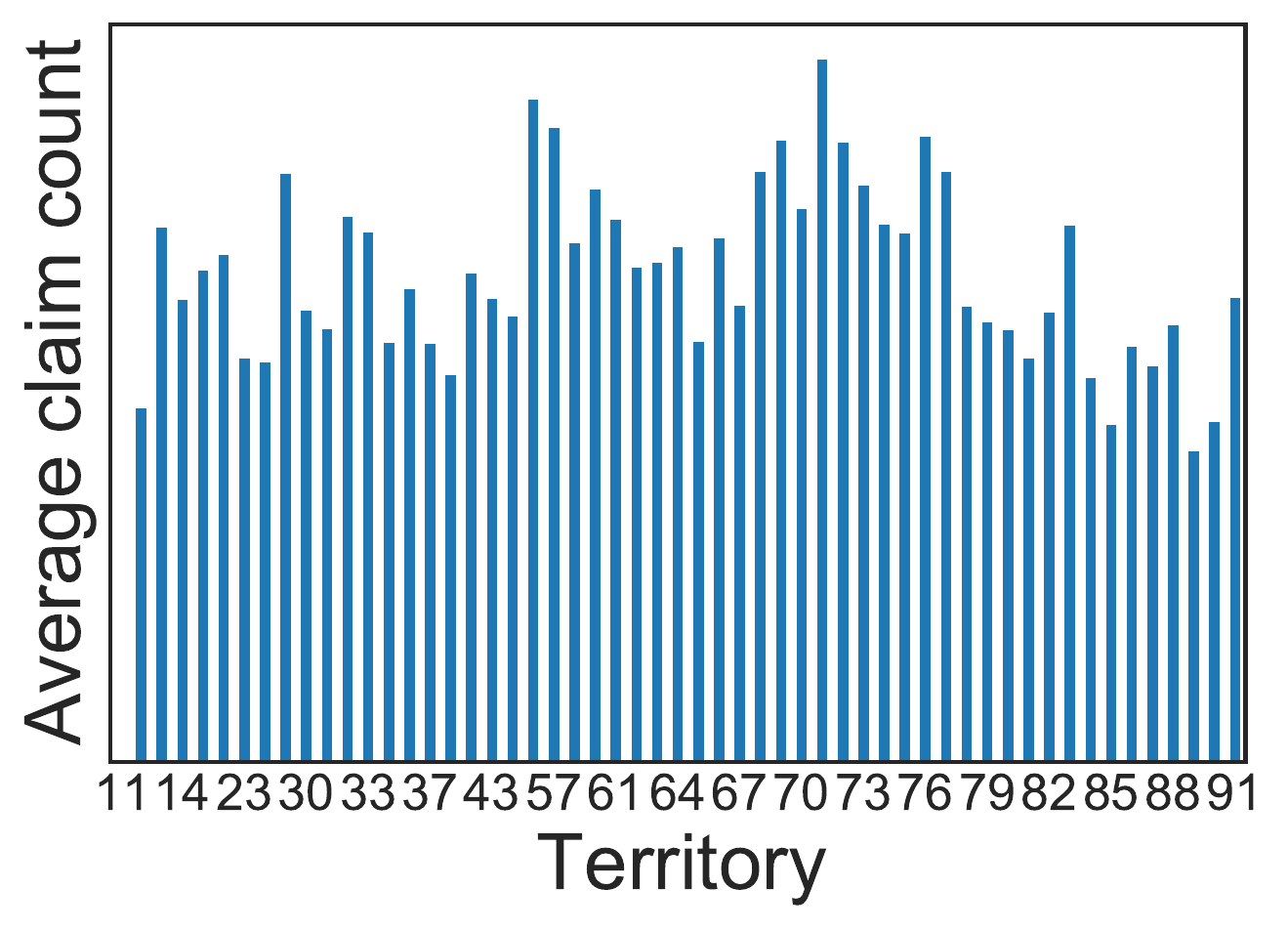} 
	\caption{Synthetic data: Distribution of average number of claims for four traditional features.} \label{fig11:dist_trad_syn}
\end{figure}

\begin{figure}[htbp]
	\centering
	\includegraphics[scale=0.3]{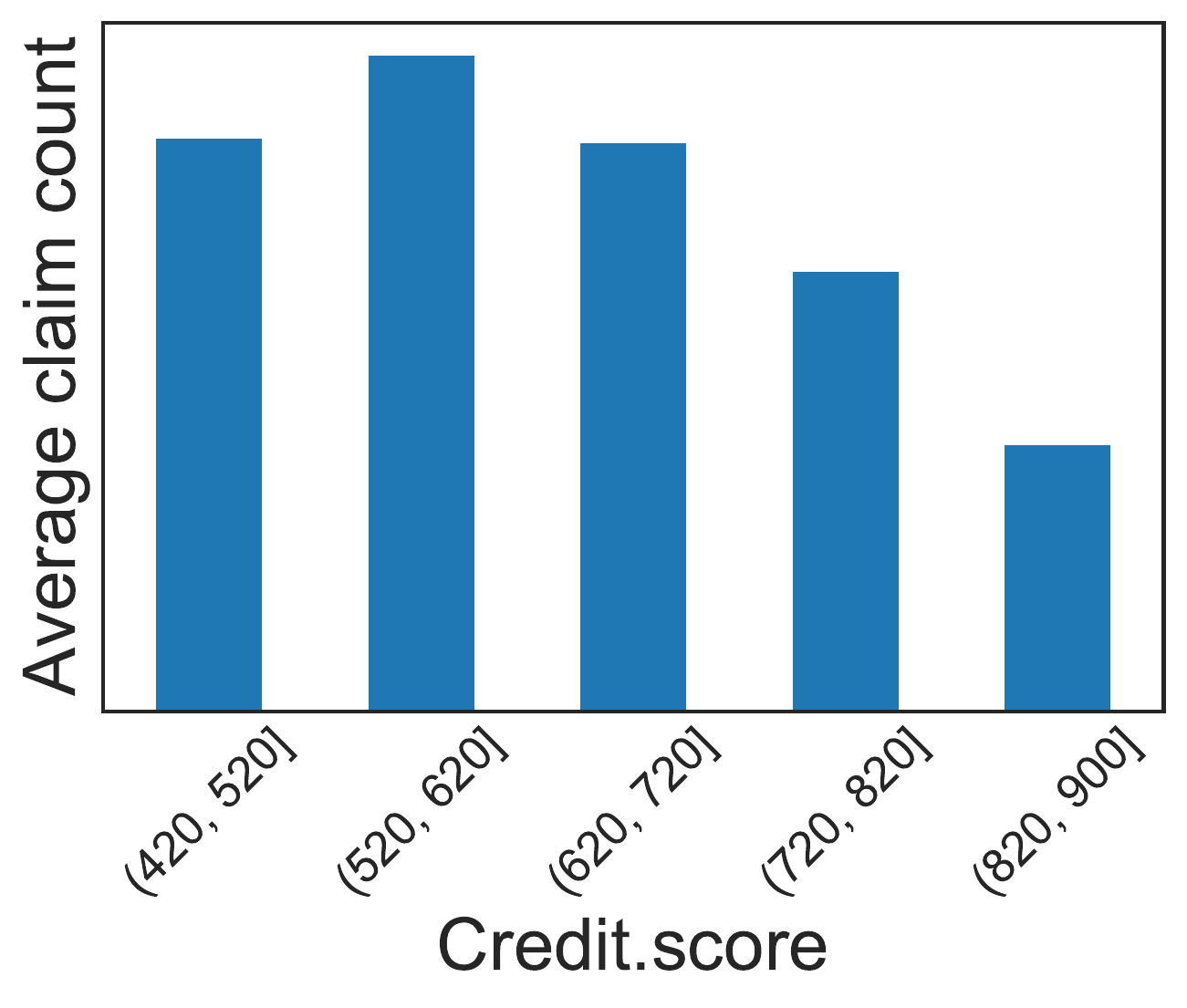} 
	\includegraphics[scale=0.3]{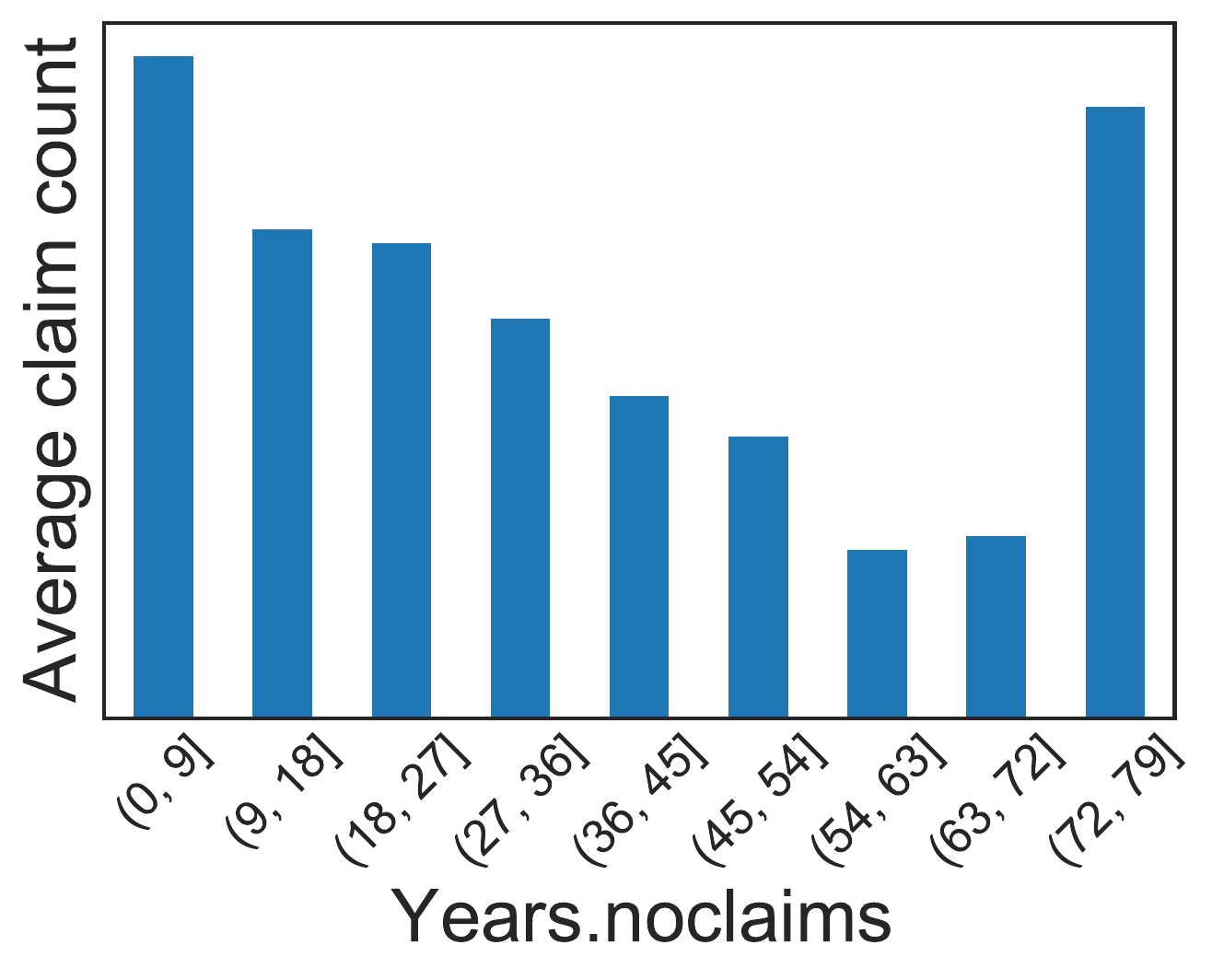} 
	\includegraphics[scale=0.3]{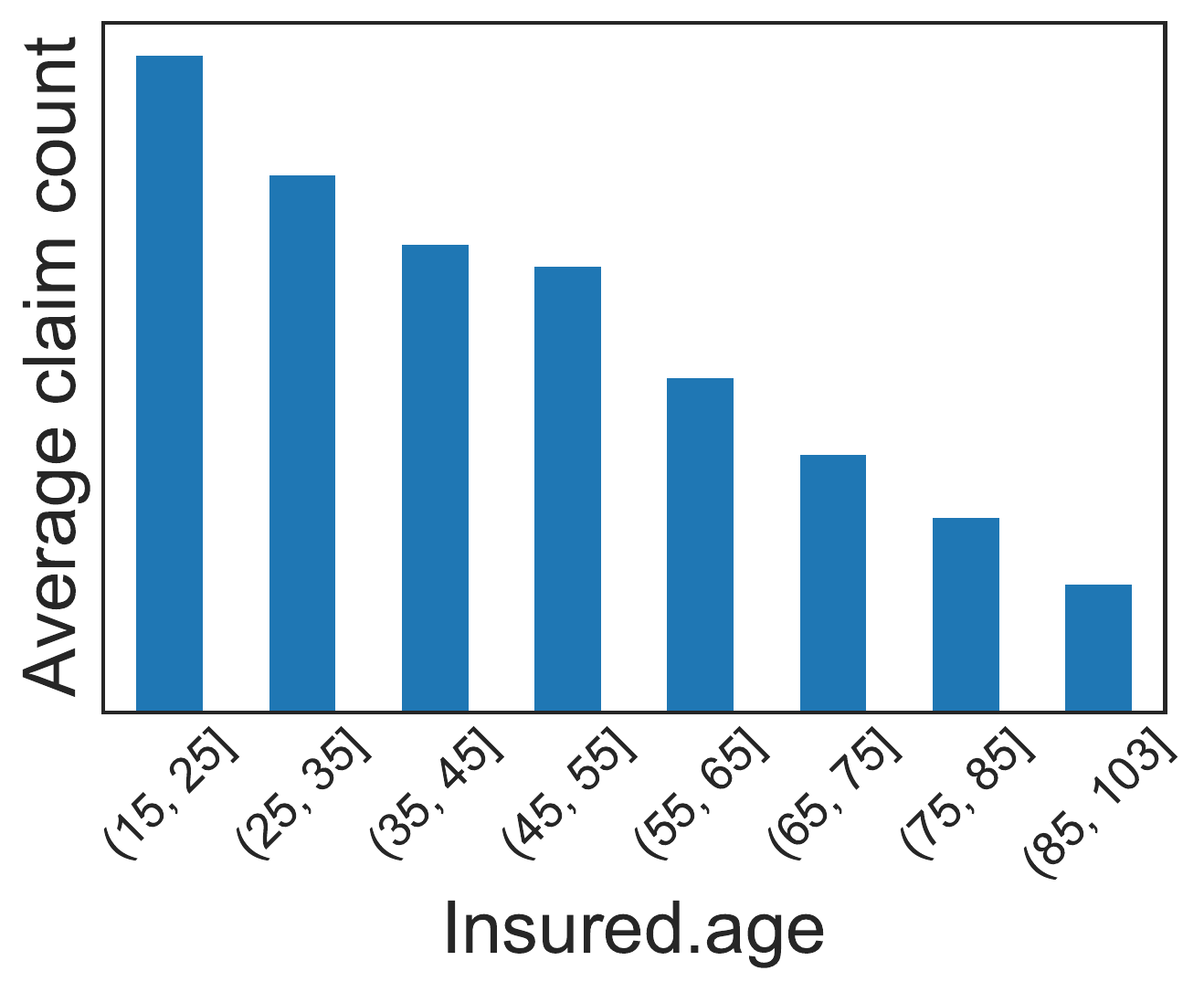} 
	\includegraphics[scale=0.3]{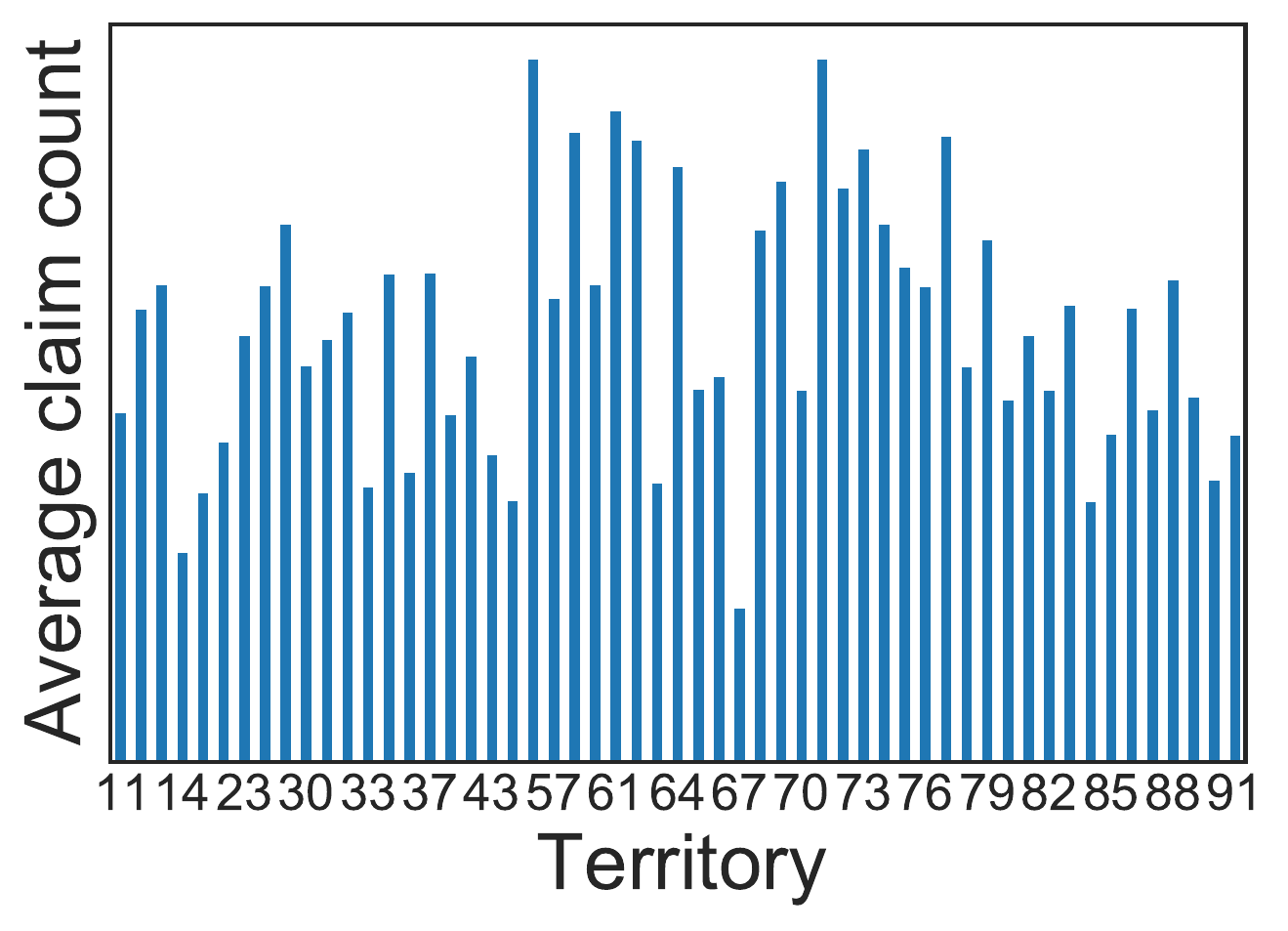} 
	\caption{Real data: Distribution of average number of claims for four traditional features.} \label{fig12:dist_trad_real}
\end{figure}

\bigskip
\newpage

\subsection*{Acknowledgements} \label{sec:ty}

The authors are very thankful for the financial support provided by the CAE (Centers of Actuarial Excellence) research grant on \textit{Applying Data Mining Techniques in Actuarial Science} from the Society of Actuaries (SOA). Banghee So acknowledges and thanks the support from the SOA through its James C. Hickman Scholar program.

\setlength{\bibsep}{0.7ex plus 1ex}
\bibliographystyle{apalike}
\bibliography{tel_syn_v4.bib}

\end{document}